\documentclass[pdflatex,sn-mathphys-num, iicol]{sn-jnl}

\usepackage{graphicx}%
\usepackage{multirow}%
\usepackage{amsmath,amssymb,amsfonts}%
\usepackage{amsthm}%
\usepackage{mathrsfs}%
\usepackage[title]{appendix}%
\usepackage{xcolor}%
\usepackage{textcomp}%
\usepackage{manyfoot}%
\usepackage{booktabs}%
\usepackage{algorithm}%
\usepackage{algorithmicx}%
\usepackage{algpseudocode}%
\usepackage{listings}%

\usepackage{orcidlink}
\usepackage{bbding}
\usepackage{rotating}
\usepackage{tabularx}
\usepackage{tabularray}
\usepackage{xspace}
\usepackage{capt-of}
\usepackage{pifont}
\usepackage{ulem}

\newcommand{\ie}{\textit{i}.\textit{e}.,\xspace}
\newcommand{\eg}{\textit{e}.\textit{g}.,\xspace}
\newcommand{\APBEV}{AP$_\text{BEV}$\xspace}

\newcommand{\APthreeD}{AP$_\text{3D}$\xspace}
\newcommand{\APnew}{AP$_\text{CS-BEV}$\xspace}
\newcommand{\APcs}{AP$_\text{CS-ABS}$\xspace}

\theoremstyle{thmstyleone}%
\theoremstyle{thmstyletwo}%

\theoremstyle{thmstylethree}%

\begin{document}

\title[CornerPoint3D: Look at the Nearest Corner Instead of the Center]{CornerPoint3D: Look at the Nearest Corner Instead of the Center}


\author*[1]{\fnm{Ruixiao} \sur{Zhang} \orcidlink{0000-0002-4959-831X}}\email{rz6u20@soton.ac.uk}
\equalcont{These authors contributed equally to this work.}

\author[2,3]{\fnm{Runwei} \sur{Guan}}\email{runwei.guan@liverpool.ac.uk}
\equalcont{These authors contributed equally to this work.}

\author[4]{\fnm{Xiangyu} \sur{Chen}}\email{xc429@cornell.edu}

\author[1]{\fnm{Adam} \sur{Prugel-Bennett}}\email{apb@ecs.soton.ac.uk}

\author*[1]{\fnm{Xiaohao} \sur{Cai}}\email{x.cai@soton.ac.uk}

\affil*[1]{\orgname{University of Southampton}, \orgaddress{\city{Southampton}, \postcode{SO17 1BJ}, \country{UK}}}

\affil[2]{\orgdiv{Thrust of Artifial Intelligence}, \orgname{Hong Kong University of Science and Technology (Guangzhou)}, \city{Guangzhou}, \postcode{511453}, \country{China}}

\affil[3]{\orgdiv{Department of Electrical Engineering and Electronics}, \orgname{University of Liverpool}, \city{Liverpool}, \postcode{L69 3BX}, \country{UK}}

\affil[4]{\orgdiv{Department of Computer Science}, \orgname{Cornell University}, \orgaddress{\street{107 Hoy Rd}, \city{Ithaca}, \postcode{14850}, \state{New York}, \country{USA}}}


\abstract{3D object detection aims to predict object centers, dimensions, and rotations from LiDAR point clouds. Despite its simplicity, this is inherently an ill-posed problem: LiDAR captures only the near side of objects, making center-based detectors prone to poor localization accuracy in cross-domain tasks with varying point distributions. Meanwhile, existing evaluation metrics designed for single-domain assessment also suffer from overfitting due to dataset-specific size variations. A key question arises: Do we really need models to maintain excellent performance in the entire 3D bounding boxes after being applied across domains? From a practical application perspective, one of our main focuses is on preventing collisions between vehicles and other obstacles, especially in cross-domain scenarios where correctly predicting the sizes is much more difficult. To address these issues, we rethink cross-domain 3D object detection from a practical perspective. We propose two new metrics that evaluate a model’s ability to detect objects' closer-surfaces to the LiDAR sensor, providing a more comprehensive cross-domain assessment. Additionally, we introduce EdgeHead, a refinement head that guides models to focus more on learnable closer surfaces, significantly improving cross-domain performance under both our new and traditional BEV/3D metrics. Furthermore, we argue that predicting the nearest corner rather than the object center enhances robustness. We propose a novel 3D object detector, coined as CornerPoint3D, which is built upon the representative detector CenterPoint, and uses heatmaps to supervise the learning and detection of the nearest corner of each object instead of the center. Our proposed methods realize a balanced trade-off between the detection quality of entire bounding boxes and the locating accuracy of closer surfaces to the LiDAR sensor, outperforming the traditional center-based detector CenterPoint in multiple cross-domain tasks and providing a more practically reasonable and robust cross-domain 3D object detection solution.}


\maketitle

\section{Introduction}

3D object detection aims to localize and categorize different types of objects in specific 3D space described by 3D sensor data (\eg LiDAR point clouds). Recently, the application of this technology has achieved significant improvement due to the development of deep neural networks, especially in the field of autonomous driving. Current 3D object detection methods mainly focus on specific datasets, \ie models will be trained and tested independently on a specific dataset. In doing so, a number of models achieved high performances on public benchmarks including nuScenes~\cite{9156412}, Waymo~\cite{Sun_2020_CVPR}, and KITTI~\cite{6248074}.  However, if the application of the model on a new dataset is needed, the training on the new dataset as well as modifications of some training hyper-parameters are usually necessary. In other words, it is hard for models trained on one dataset to adapt directly to another. These domain shifts may arise from different sensor types, weather conditions~\cite{9156543}, and object sizes~\cite{9578132} between different datasets or domains. This cross-domain problem is therefore a big challenge for real-world applications of existing 3D object detection methods, as their retraining steps can be very slow and resource-consuming. It is thus significant to understand the reasons for the drop in cross-domain performance and propose efficient methods to raise the cross-domain performance to the same level as within-domain tasks.

\begin{figure}[!htb]
\begin{center}
   \includegraphics[width=0.98\linewidth]{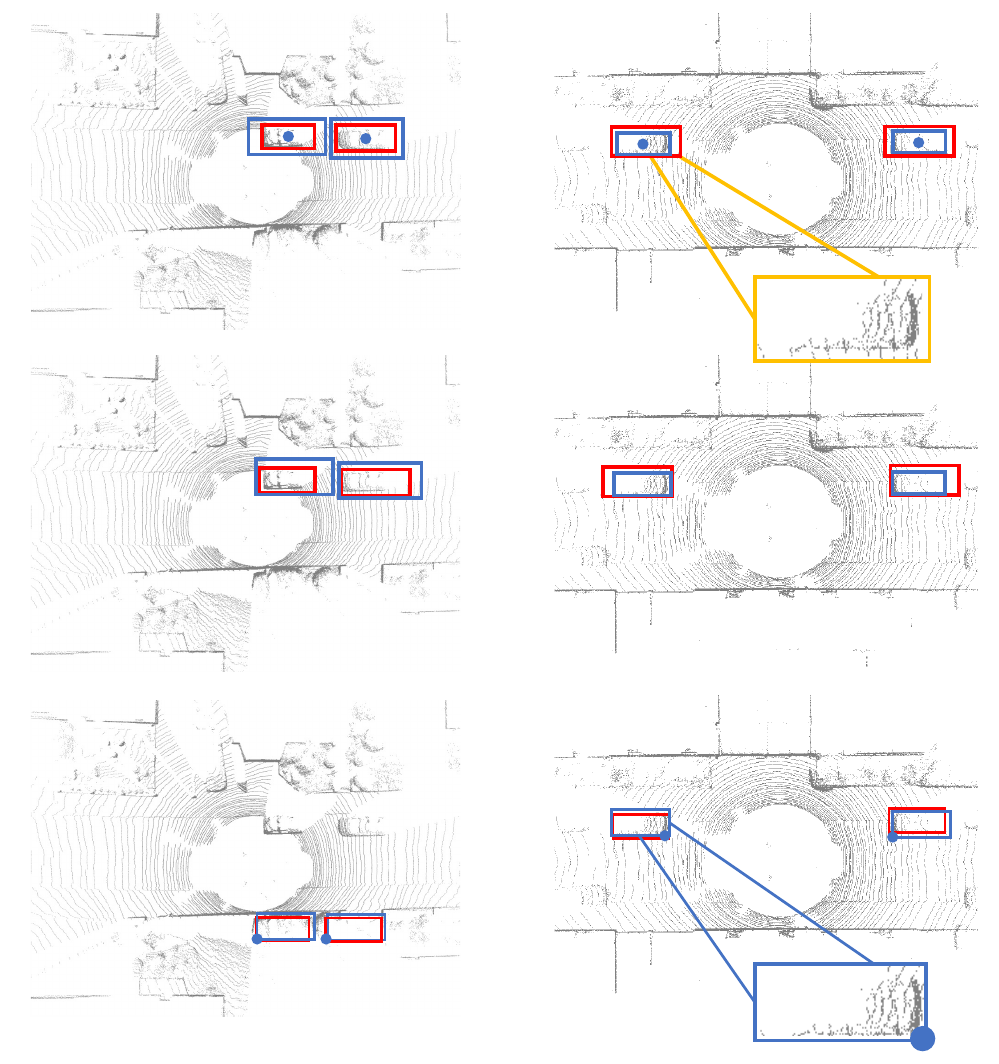}
   \put(-205,150){\footnotesize (a) Traditional center-based methods}
   \put(-205,77){\footnotesize (b) Methods equipped with our EdgeHead}
   \put(-205,4){\footnotesize (c) Our CornerPoint3D detector}
\end{center}
   \caption[CornerPoint3D]{Prediction illustrations of different detectors in cross-domain tasks. The left and right columns showcase the prediction properties when the training domain respectively has a larger and smaller average object size than the target domain. (a) Traditional center-based methods predict the object centers without available point cloud data surrounding them, often resulting in overfitting and guessing about the center location based on average object sizes in training data. (b) Our proposed refinement head, EdgeHead, improves the detection of surfaces closer to the ego vehicle, mitigating the impact of size overfitting in occlusion avoidance. (c) Our proposed novel detector based on the nearest corner prediction, enabling more robust detection in cross-domain 3D object detection tasks. }
\label{fig:figure1}
\end{figure}


One of the core challenges in cross-domain 3D object detection is the overfitting of models on the training domains. As discussed in existing works~\cite{wang2020train, zhang2024detect, zhang2024revisiting}, models tend to make predictions with similar size distributions to the training domain, even if they are applied to target domains with totally different object size distributions. In fact, existing models normally predict variables that are not directly related to available information from the point cloud data. Both the anchor-based and anchor-free models utilize a center-based representation of the object bounding boxes, which heavily relies on the accuracy of the center prediction. However, the information relative to the center is not directly included in the original point cloud data as LiDAR sensors mainly capture points surrounding the closer surfaces of objects toward the ego vehicle. Meanwhile, as shown in Figure~\ref{fig:figure1}, many objects' point clouds are incomplete due to physical obstructions in the capturing process of LiDAR sensors. As a result, such center predictions sometimes cannot get enough knowledge from the data itself but have to overfit to the ground-truth annotations in the training domain, leading to bigger errors when further applied to other domains.

From another point of view, existing metrics are usually designed to evaluate models' ability to predict the complete shapes of objects. However, since most objects' point clouds are incomplete due to physical obstructions, it is difficult to predict the full size and entire box of objects. In within-domain tasks, we may be able to ``guess" the prediction through training with large amounts of samples. However, the guessing easily becomes incorrect in cross-domain tasks since the average object size has changed. Therefore, correctly predicting the location (usually represented by the center of the box) along with the full box size becomes a much more difficult task. Instead, we argue that what we can better guide the model to learn from the incomplete object point clouds is to predict the closer surfaces of the objects to the ego vehicle, since there are plenty of points over there as shown in Figure~\ref{fig:figure1}. Among the four surfaces perpendicular to the ground, capturing and describing these two is more reasonable and also more important, since our main purpose is to avoid potential collisions with surrounding objects during driving.

Some generic training methods have tried to overcome the cross-domain problem. Among these, some methods explore the possibility of adding information about domain gaps into the models by training or using additional prior knowledge to adapt the trained models from the training domain to new domains~\citep{9156543, 9578132, yuan2023bi3d, chen2023revisiting}. Some other methods tried to improve the models' generalization ability to enhance their performance on multiple domains without fine-tuning and additional training operations~\citep{hu2023density, zhang2023uni3d}. However, most of these approaches still prioritize the detection quality of entire bounding boxes and retain the existing model design strategy (\ie still centered on object locations), limiting their ability to fundamentally address the overfitting problem.

In this paper, we approach the problem of cross-domain 3D object detection from a different perspective, showing how properly using the knowledge from the original data can greatly improve the models' 3D object detection performance, especially in cross-domain tasks.

Firstly, we propose two metrics, named \APnew and \APcs, to measure 3D object detection models' ability of detecting the closer surfaces to the sensor on the ego vehicle, which can be used with the commonly used metrics \APBEV and \APthreeD, \ie bird's-eye view (BEV) and 3D average precision (AP), to evaluate existing models' cross-domain performance more comprehensively and reasonably.

We then propose the EdgeHead, a second-stage refinement head that can be applied to existing models and guide them to focus more on the learnable closer surfaces of objects. By applying the EdgeHead to two representative detectors CenterPoint~\citep{Yin_2021_CVPR} and SECOND~\citep{s18103337} under multiple cross-domain tasks, we show that EdgeHead can effectively help models detect better the closer surfaces and perform better under both the existing metrics and our proposed metrics. Extensive experiment results indicate that by guiding models to focus more on the surfaces with more points captured by the LiDAR sensor, the models can learn more robust knowledge from the training domain and perform better in cross-domain tasks.

Furthermore, we propose a novel 3D object detector named CornerPoint3D, which focuses more on the nearest corners of objects towards the LiDAR sensor and the ego vehicle. Specifically, CornerPoint3D is built upon the representative anchor-free detector, CenterPoint~\citep{Yin_2021_CVPR}. It uses a standard LiDAR-based 3D backbone network to extract 3D features from point cloud data and flatten them into 2D BEV features. Afterwards, the objects' nearest corners to the LiDAR sensor are predicted via a heatmap-based module. For each detected corner, we regress all other object properties related to it, including the box dimensions and rotations. To reduce the potential false positive bounding boxes that arise from the corner-based box generation process, an additional separate head is also utilized to predict the relative position vector between the nearest corner and the center of the bounding boxes, providing extra guidance on the selection of the bounding box according to the corners. More details are discussed in Section~\ref{sec:cornerpoint3d-separatehead}. As introduced in Section~\ref{sec:cornerpoint3d-multigated}, a {\it multi-scale gated module} (MSGM) is also designed for adaptive feature extraction in cross-domain tasks. When compared with the CenterPoint detector, our CornerPoint3D achieves better performance in cross-domain tasks under the \APnew and \APcs metrics for closer-surfaces detection ability.

With both methods focusing on the closer surfaces of objects, where more prior knowledge (\ie point cloud data) is available, we further apply the EdgeHead as the second-stage head of our CornerPoint3D, which is named as CornerPoint3D-Edge. In multiple cross-domain tasks, CornerPoint3D-Edge surpasses the CenterPoint equipped with EdgeHead under all four evaluation metrics (\ie \APBEV, \APthreeD, \APnew, and \APcs). Finally, with the application of the data augmentation method ROS~\cite{9578132}, CornerPoint3D-Edge achieves the state-of-the-art (SOTA) performance, demonstrating robust and impressive detection ability in cross-domain 3D object detection tasks, and strong compatibility with other cross-domain data augmentation and generalization methods.

This work presents a significant extension compared to our previous work~\cite{zhang2024detect}, specifically in the development of the nearest corner-based detector CornerPoint3D. In the conference version, we proposed the EdgeHead as the first (to our knowledge) to explore the possible effect of guiding models to focus more on the parts of objects where more points are available. With the proposed additional evaluation metrics, we prove that such a strategy is workable and worth further exploration. However, EdgeHead is just a refinement head and has to be equipped with existing models that still focus on entire objects, which limits the concentration on the richer point cloud data surrounding their closer surfaces. Therefore, in this work, we further propose a novel and independent detector, \ie CornerPoint3D. By focusing more on the nearest corners of objects towards the LiDAR sensor, CornerPoint3D realizes the concentration on visible point cloud data at the early stages, beginning at the generation of the nearest corner heatmaps just after the backbones. With CornerPoint3D, we are now able to apply the EdgeHead to a detector with the same focus, fully unlocking the potential of both methods and realizing a balanced trade-off between the detection quality of the entire bounding boxes and the locating accuracy of closer surfaces to the LiDAR sensor. With further cooperation with the data augmentation method ROS~\citep{9578132}, we finally achieve a from-start-to-end 3D object detection system that aims to improve the generalization ability by focusing on the available prior knowledge (\ie the point cloud data), providing a more practically reasonable and robust 3D object detection solution.

In sum, our contributions are four-fold as follows.

(1) We propose two additional evaluation metrics, the \APnew and \APcs, to measure 3D object detection models' ability to detect the closer surfaces to the sensor on the ego vehicle, which can be used with the commonly applied metrics \APBEV and \APthreeD to evaluate model's cross-domain performance more comprehensively and reasonably.

(2) We propose the EdgeHead, a second-stage refinement head to guide models to focus more on the learnable closer surfaces of objects, which can be easily equipped to existing models and greatly improve their cross-domain performance. We apply the EdgeHead to two representative 3D detectors, the CenterPoint~\citep{Yin_2021_CVPR} and the SECOND~\cite{s18103337}, consistently achieving improvements.

(3) We propose the CornerPoint3D detector with the nearest corner-based representation of 3D object bounding boxes and the MSGM, guiding the model to focus more on the nearest corner and the closer surfaces of objects, and improving the robustness in cross-domain tasks. 

(4) We combine our proposed CornerPoint3D and EdgeHead and evaluate them on three challenging cross-domain tasks, achieving significantly improved performance compared to the prior arts. Our method realizes a balanced trade-off between the detection quality of the entire bounding boxes and the locating accuracy of closer surfaces to the LiDAR sensor, providing a more practically reasonable and robust 3D object detection solution.

\section{Related work}

\subsection{3D object detection with point clouds} LiDAR point cloud is a commonly used representation of real-world 3D spaces. Benefiting from accurate point locations, the LiDAR point cloud has been widely used in 3D object detection tasks, especially in autonomous driving systems. Existing LiDAR-based methods can be roughly divided into two main categories, \ie the voxel-based methods and the point-based methods, according to their ways of processing the point cloud data. Voxel-based methods like VoxelNet~\cite{8578570} and SECOND~\cite{s18103337} utilized a voxel-based representation for the point clouds and then used 3D CNNs to extract features from the voxels. PointPillars~\citep{lang2019pointpillars} used pillars, a special type of voxel, to group the points, achieving significantly higher speed than contemporary voxel-based methods while maintaining comparable performance. Point-based methods, such as PointRCNN~\cite{shi2019pointrcnn}, applied PointNet++~\cite{qi2017pointnet++} from semantic segmentation to obtain 3D point-wise features and directly learned the 3D proposals from the points. 3DSSD~\citep{9156597} proposed a fusion sampling strategy in the feature space, further improving the detection quality and speed of PointRCNN. PV-RCNN~\cite{9157234} combined the voxel-based and point-based methods, which voxelized the point cloud data first and involved more semantic features through key-point-wise feature extraction, greatly improving the performance with acceptable computation cost. More recently, some researchers also attempted to use transformer-based~\cite{dosovitskiy2020vit, carion2020end} structures in 3D object detection. For example, TransFusion~\cite{Bai_2022_CVPR} used the transformer decoder to combine image features and LiDAR point cloud features through a soft association for prediction quality enhancement.

\subsection{Domain adaptation} 

Domain adaptation has been widely applied in 2D object detection~\cite{Hsu_2020_WACV, 9008383, 8953674} to bridge the gaps between different datasets and tasks. \citet{chen2018domain} performed feature alignment at both image-level and instance-level upon the Faster R-CNN detector. \citet{lin2021domain} used a domain-invariant network with representation reconstruction to learn disentangled representations. SWDA~\citep{saito2019strong} proposed the strong-weak distribution alignment for local features rather than for global features. D-adapt \citep{jiang2021decoupled} proposed the decoupled adaptation method to decouple the adversarial adaptation and the training of the detector. \citet{zhang2022towards} proposed a novel method named region aware proposal reweighting to eliminate dependence within region of interest (RoI) features, outperforming other 2D object detection methods in domain generalization. However, only a few domain adaptation approaches are specifically designed for 3D object detection~\cite{wang2020train, 9578132, yuan2023bi3d}. Standard normalization (SN)~\cite{9156543} comprehensively analyzed the overfitting problem in object sizes for cross-domain tasks.  By normalizing the object size of different datasets, it proposed a simple but effective solution. ST3D~\cite{9578132} and ST3D++~\citep{yang2021st3d++} introduced the random object scaling (ROS) algorithm to enhance model robustness in car size prediction and applied self-training strategies to generate pseudo labels, enabling training on unseen data without ground truth. \citet{wei2022lidar} proposed a distillation method for LiDAR point clouds to overcome the beam difference between datasets, which enables models to adapt to point cloud data with lower densities and fewer beams.

\subsection{Domain generalization} There are also explorations~\cite{zhang2023uni3d, hu2023density} on improving models' generalization ability, \ie training models only once and directly applying them to multiple domains. In 2D object detection, early works focused on merging the taxonomy information and training the models on a unified label space. \citet{zhao2020object} designed a framework which works with partial annotations that are annotated in one dataset but not in another, and exploited a pseudo-labeling approach for these specific cases. More recently, \citet{dai2021dynamic} proposed a novel dynamic head framework to unify object detection heads with attention, significantly improving the representation ability of object detection heads. \citet{wang2019towards} proposed a domain adaptation layer and attention mechanism to alleviate the dataset-level differences. \citet{zhou2022simple} introduced a novel automatic way to merge the taxonomy space without the need of manual taxonomy reconciliation. In 3D object detection, a multi-domain knowledge transfer framework was introduced in \cite{wu2023towards}, utilizing spatial-wise and channel-wise knowledge sharing across domains, enabling the extraction of universal feature representations. The random beam re-sampling method was proposed in \cite{hu2023density} to improve the models' beam-density robustness and used a teacher-student framework to generate pseudo labels on unseen target domains. \citet{zhang2024detect} proposed EdgeHead, a refinement head that guides models to focus on the closer surfaces of objects where more reliable point information is available, thereby enhancing the cross-domain performance of existing models.

\section{Evaluation for closer-surfaces detection} \label{sec:metricdesign}

We first point out the weakness of existing metrics such as BEV and 3D AP in cross-domain tasks. Given two predictions of the same car ground truth as shown in Figure~\ref{fig:figure1}(a) and Figure~\ref{fig:figure1}(b), exactly the same BEV and 3D AP will be obtained as they have the same overlaps with the ground-truth bounding box. However, there is a big difference when comparing their detection quality of the surfaces that are closer to the LiDAR sensor, which are not occluded by other surfaces and therefore have a bigger chance of being captured by the sensor. This detection quality indicates how correctly a model can estimate the distance from our car to the surfaces of other objects that could collide with us, which is directly related to driving safety and should be paid more attention to than the other two surfaces of detected objects. It is therefore essential to develop additional metrics that can accurately assess this detection quality of models, thereby enabling a more reasonable and comprehensive evaluation of performance, particularly for cross-domain tasks. Therefore, we aim to measure models' cross-domain 3D object detection performance with fairer and more reasonable metrics. Specifically, we propose two evaluation metrics that are influenced by the cross-domain factors (\eg object sizes and point cloud densities).

\subsection{Metrics \APcs and \APnew} We first define the absolute gap between the closer surfaces of predictions and ground truth. Given a prediction box with its vertices $\left\{V_{\rm pred}^i \right\} _{i=1}^4$ on the BEV plane and the related ground-truth box with its vertices $\left\{V_{\rm gt}^i \right\} _{i=1}^4$, we first sort their vertices by their distance to the origin (\ie the location of the LiDAR sensor), and then further sort the second and third vertices by their absolute x-coordinate. After sorting, prediction and ground-truth boxes should follow the same indexing rule for their vertices, \ie $V^1$ and $V^4$ are respectively the vertices closest and furthest to the origin, and $V^2$ is the vertex having a smaller absolute x-coordinate compared with $V^3$. We can then define the absolute gap, say $G_{\rm cs}$, of the closer surfaces between the prediction and the ground truth, \ie
\begin{align} \label{eq:g-cs}
    \begin{split}
	G_{\rm cs} = & \ | V_{\rm pred}^1 - V_{\rm gt}^1| + {\rm Dist}(V_{\rm pred}^2, E_{\rm gt}^{1,2}) \\
    & \ + {\rm Dist}(V_{\rm pred}^3, E_{\rm gt}^{1,3}),
    \end{split}
\end{align}
where $E^{i,j}$ is the edge connecting $V^i$ and $V^j$, and ${\rm Dist}(V,E)$ calculates the perpendicular distance from vertex $V$ to edge $E$.

The defined absolute gap $G_{\rm cs}$ in Eq. \eqref{eq:g-cs} can be used to measure the detection quality of the closer surfaces; however, it will fluctuate with the sizes of the object boxes. In other words, it is not a scaled metric that can be used to calculate the AP with pre-determined thresholds. 
To solve this problem, we propose the \textbf{\textit{absolute closer-surfaces AP}} (\ie the \APcs) to directly measure the detection quality of the closer surfaces by using
\begin{align} \label{eq:abs-cs}
    \Gamma_{\rm ABS}^{\rm CS} = {1}/{(1+\alpha  G_{\rm cs})},
\end{align}
where $\alpha\ge 0$ is the penalty ratio set to 1 by default. 

The proposed CS-ABS AP by Eq. \eqref{eq:abs-cs} can also be utilized to combine with existing popular metrics and thus form new metrics with hybrid effectiveness for more powerful and fairer evaluation of models' performance.
In particular, we combine the CS-ABS AP with the BEV AP and propose the \textbf{\textit{closer-surfaces penalized BEV AP}} (\ie the \APnew) to measure the detection quality by using the penalized IoU, say $\Gamma_{\rm BEV}^{\rm CS}$, \ie
\begin{align} \label{eq:bev-cs}
    \Gamma_{\rm BEV}^{\rm CS }= {\Gamma_{\rm BEV}}/{(1+\alpha  G_{\rm cs})}
\end{align}
where $\Gamma_{\rm BEV}$ is the original BEV IoU and $\alpha$ is the penalty ratio set to 1 by default based on experimental experience (see more discussions in Section~\ref{sec:ablation_corner3d}). 
%
Our proposed \APnew metric in Eq. \eqref{eq:bev-cs} not only retains the robustness of the original BEV metric but also better distinguishes the detection quality of the closer surfaces. It finds an evaluation balance between the quality of the entire 3D box and the quality of the closer surfaces. Taking the same examples in Figure~\ref{fig:figure1}(b), the newly proposed metric will return a higher AP when the prediction matches the closer surfaces of the ground truth better.

In sum, we in this section proposed two evaluation metrics: \APcs and \APnew. The \APcs can directly tell the detection quality of the closer surfaces without considering the ability to evaluate the quality of the entire 3D box, which can be specifically used when analyzing the detection quality gain regarding closer surfaces. The \APnew 
finds a balance between the quality of the entire 3D box and the closer surfaces, which is more comprehensive and can be used to measure the overall cross-domain performance for different models and tasks.

\section{EdgeHead}

Existing models are easy to overfit on the training domain, on which they are designed and trained to perform well. This limits their detection ability across domains. One of the main factors that causes this overfitting problem is that these models usually have a regression module to learn the offsets of box dimensions and locations between the prediction and ground truth. Such a module results in the overfitting on object sizes, especially for anchor-based methods. It is thus critical to explore the performance of a specifically designed model with the consistent aim of the new metric we usher in here, \ie the one that focuses more on the detection quality of the closer surfaces of the bounding box. 

In this section, we briefly review our proposed EdgeHead~\citep{zhang2024detect}, a second-stage refinement head to guide models to focus more on the learnable closer surfaces of objects, which can be easily equipped to existing models and greatly improve their cross-domain performance.

\begin{figure}[t]
\centering
   \includegraphics[width=0.95\linewidth]{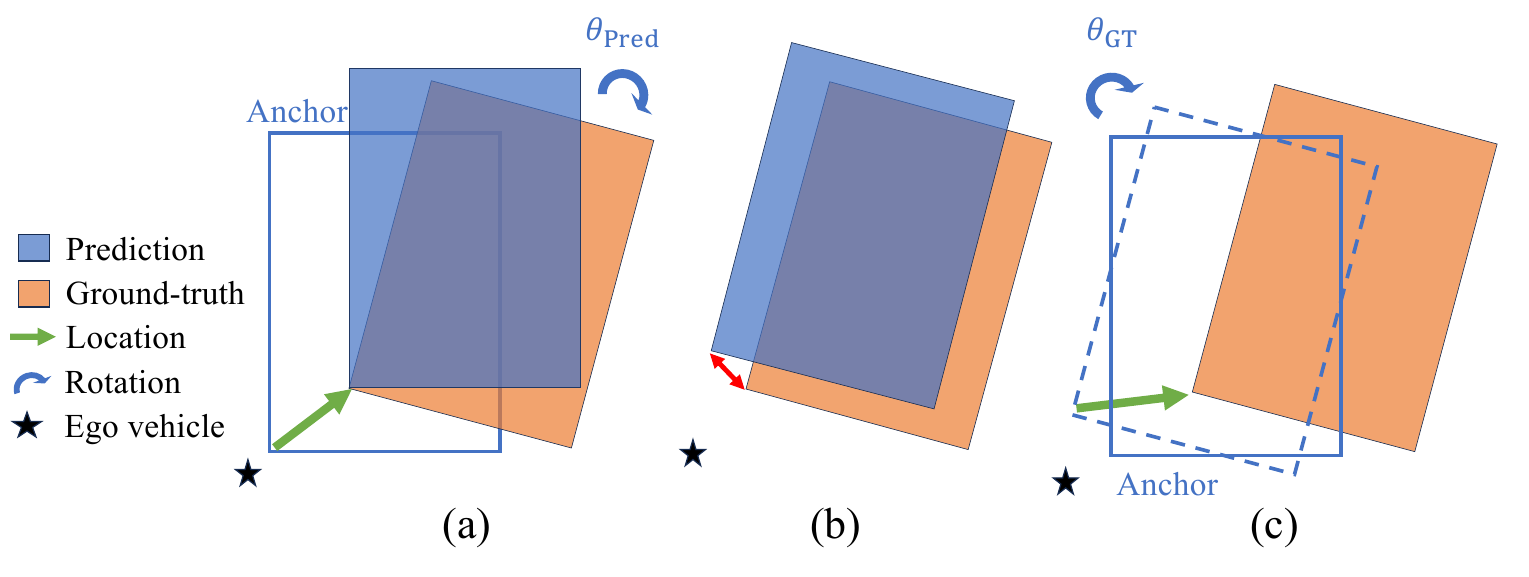}
\caption{Illustration of different regression processes. (a) The process that directly regresses the closest vertex and rotations without rotating the anchor box first. (b) The prediction obtained using the process in (a), in which the red arrow shows that the prediction does not learn the closest vertex as expected. (c) The regression process guided by Eq.~\eqref{eq:final_reg_loss} and Eq.~\eqref{eq:reg_loss_new} in our EdgeHead, which first rotates the anchor by $\theta_{\rm gt}$ and then calculates the regression target of $x$ and $y$ locations.}
\label{fig:long}
\label{fig:regress_corner}
\end{figure}

\subsection{EdgeHead for closer-surfaces localization} The main purpose of EdgeHead is to modify the models' training purpose and guide them to focus more on the closer surfaces of objects instead of the entire boxes. Similar to other refinement heads, the proposed EdgeHead takes the predictions of a model's first stage as the RoIs. It then aggregates the features from earlier backbones of the model (\eg the 3D convolution backbones) into the RoI features for prediction refinement. During the refinement process of EdgeHead, we modify the loss function to guide the model to learn the closer-surfaces offsets between the predictions and the ground truth. 

The voxel RoI pooling~\citep{deng2021voxel} is used to aggregate the RoI features. In detail, we extract the 3D voxel features from the last two layers in the 3D sparse convolution backbone, which is available for most voxel-based 3D object detection models. Afterwards, the feature of each RoI is assigned by aggregating the 3D features from its neighbor voxels via the voxel query operation. Since features from the 3D backbones usually contain more spatial and structural information, the aggregated RoI features can help improve the detection quality of the closer surfaces of bounding boxes. 

The loss of a typical RoI refinement module consists of two parts, \ie the IoU-based classification loss~\citep{9157234} and the regression loss. In detail, the original regression loss, say ${\cal L}_{\rm reg}$, uses the smooth $\ell_1$ loss~\citep{7410526} to learn the 7 parameters of the bounding boxes, \ie
\begin{align}\label{eq:origin_loss}
    {\cal L}_{\rm reg} = \sum\limits_{r \in \left\{ x_{\rm c}, y_{\rm c}, z_{\rm c}, l, h, w, \theta \right\} } {\cal L}_{{\rm smooth}-\ell_1}(\widehat{\Delta r^{a}}, \Delta r^{a})
\end{align}
where $\widehat{\Delta r^{a}}$ and $\Delta r^{a}$ are the predicted residual and the regression target, respectively. In our EdgeHead, we use the original classification loss and modify the regression loss in Eq.~\eqref{eq:origin_loss}, which will guide the model to learn the closer-surfaces offsets between the predictions and ground truth.

Given the closest vertex of the anchor box and the ground-truth box respectively as $(x_{\rm cv}^{a}, y_{\rm cv}^{a}, z_{\rm cv}^{a})$ and $(x_{\rm cv}^{\rm gt}$, $y_{\rm cv}^{\rm gt}, z_{\rm cv}^{\rm gt})$, we first rotate the anchor box by the rotation angle $\theta_{\rm gt}$ of the ground-truth box as shown in Figure~\ref{fig:regress_corner}(c), and denote the rotated box's closest vertex by $(x_{\rm cv}^{a^{\prime}}, y_{\rm cv}^{a^{\prime}}, z_{\rm cv}^{a^{\prime}})$. We then calculate the residuals of $x_{\rm cv}$ and $y_{\rm cv}$ between the rotated anchor box and ground truth as follows
\begin{equation}\label{eq:final_reg_loss}
\begin{aligned}
    {\Delta x_{\rm cv}} = x_{\rm cv}^{\rm gt} - x_{\rm cv}^{a^{\prime}}, \quad
    {\Delta y_{\rm cv}} = y_{\rm cv}^{\rm gt} - y_{\rm cv}^{a^{\prime}},
\end{aligned}
\end{equation}
and modify the $\ell_1$ loss by replacing the residual of center locations with the residuals of the rotated closest vertex to the origin (\ie our ego vehicle) as calculated in Eq.~\eqref{eq:final_reg_loss}. Since the Z-axis of bounding boxes is always set to be perpendicular to the horizontal plane, the distances of the closer surfaces between the predictions and ground truth are only related to the X and Y coordinates. We therefore remove the regression for $z_{\rm cv}$ and focus on $x_{\rm cv}$ and $y_{\rm cv}$. To avoid the overfitting problem on object sizes, we also remove the parts of regression loss for the residuals related to object sizes (\ie $l,w,h$). As a result, we only refine $x_{\rm cv}$, $y_{\rm cv}$, and $\theta$ in our EdgeHead and keep the $z_{\rm cv}$, $l,w$, and $h$ as predicted by the model's first stage. The new regression loss of our EdgeHead is therefore defined as
\begin{align}\label{eq:reg_loss_new}
    {\cal L}_{\rm reg}^{\prime} = \sum\limits_{r \in \left\{ x_{\rm cv}, y_{\rm cv}, \theta \right\} } {\cal L}_{{\rm smooth}-\ell_1}(\widehat{\Delta r^{a}}, \Delta r^{a}).
\end{align}

\begin{figure*}[htbp]
\centering
\includegraphics[width=0.98\linewidth]{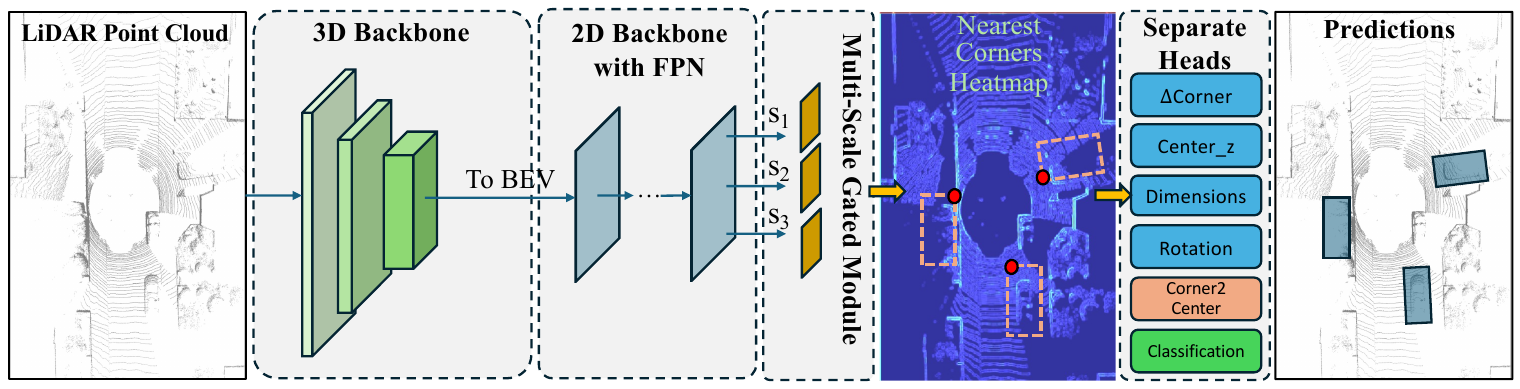}
\caption{Overview of the CornerPoint3D. Standard 3D and 2D backbones are firstly applied to obtain voxelized 3D and 2D BEV (bird’s-eye view) features from the LiDAR point cloud data. Afterwards, the BEV features are fed into the MSGM (multi-scale gated module) to extract adaptive features, which is especially essential in cross-domain 3D object detection tasks. Then, the shared multi-scale features are fed into separate 2D CNN architecture detection heads to predict the heatmaps of the objects’ nearest corners (to the LiDAR sensor) and other properties of the entire 3D bounding boxes. Afterwards, the EdgeHead~\cite{zhang2024detect} could also be utilized for second-stage refinement.} 
\label{fig:modelstructure}
\vspace{-0.05in}
\end{figure*}

\noindent \textit{Remark.}
The rotation of the anchor box used in Eq.~\eqref{eq:final_reg_loss} is important in the modification of the regression process to realize our real purpose, \ie to guide the model to learn the closer-surfaces offsets between the predictions and ground truth. Considering an example of the model's regression process without the rotation as shown in Figure~\ref{fig:regress_corner}(a), the regression target related to $x$ and $y$ will guide the model to predict the residual so that the predicted box
can coincide with the ground-truth box at the vertex closest to the origin. However, since we are also regressing the rotation angle $\theta$, we will finally get a predicted box as shown in Figure~\ref{fig:regress_corner}(b), whose closest vertex to the origin does not coincide with the ground truth's anymore (see the red arrow). To consider the rotation regression as well, we first rotate the anchor box by the rotation angle $\theta_{\rm gt}$ of the ground-truth box as shown in Figure~\ref{fig:regress_corner}(c), and then calculate the residuals of $x_{\rm cv}$ and $y_{\rm cv}$ between the rotated anchor box and ground truth as the new regression target. Such a modified regression process makes the prediction box's closest vertex finally coincide with the ground-truth box's.

\section{CornerPoint3D}

Although our proposed EdgeHead greatly improves the existing models' cross-domain performance ~\citep{zhang2024detect}, it is a refinement head and needs to be equipped with existing models that still focus on entire objects, which limits the concentration on the richer point cloud data surrounding their closer surfaces. Therefore, in this section, we further propose the novel and independent detector, \ie the CornerPoint3D. It is built upon the anchor-free detector, CenterPoint~\cite{Yin_2021_CVPR}, but focuses more on the objects' nearest corners to the LiDAR sensor. As shown in Figure~\ref{fig:modelstructure}, a standard 3D backbone is firstly applied to obtain voxelized 3D features from the LiDAR point cloud data, followed by a 2D backbone to further extract 2D BEV features. Afterwards, the BEV features are fed into the MSGM to extract adaptive features, which is especially essential in cross-domain 3D object detection tasks. Then, the shared multi-scale features are fed into separate 2D CNN architecture detection heads to predict the heatmaps of the objects' nearest corners (to the LiDAR sensor and ego vehicle) and properties of the entire 3D bounding boxes. Below we present the details of each module in CornerPoint3D.

\begin{figure}[t!]
\begin{center}
   \includegraphics[width=\linewidth]{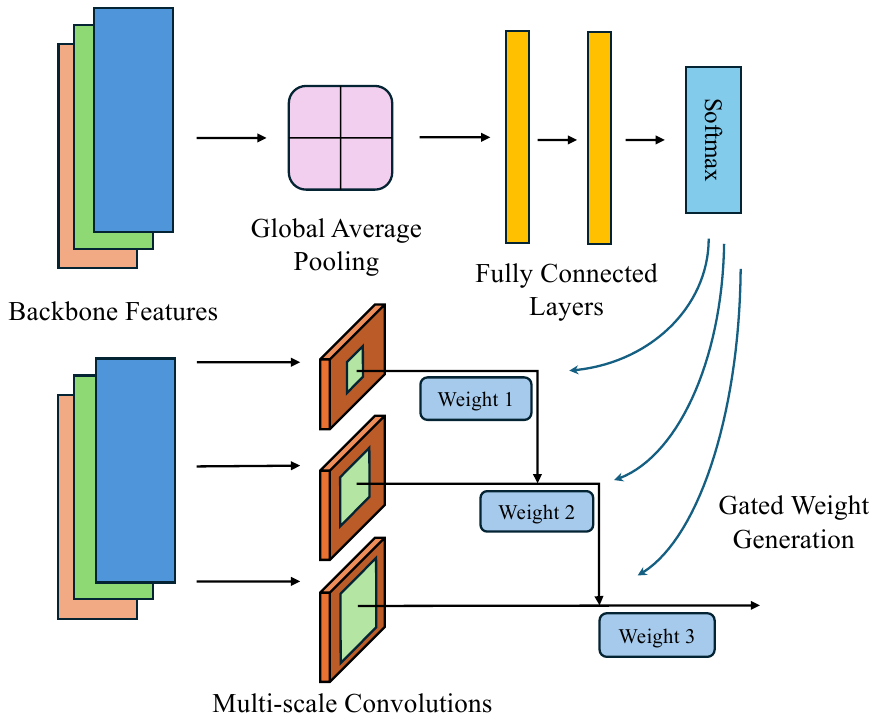}
\end{center}
    \caption[Illustration of the Multi-Scale Gated Module]{Illustration of the MSGM (multi-scale gated module). In the upper branch, backbone features are processed through a global average pooling layer, followed by two fully connected layers and a softmax function, generating gated weights for the multi-scale convolution features. In the lower branch, three convolutions with different kernel sizes are applied to capture features at multiple receptive fields. The outputs of these convolutions are then combined according to the corresponding gated weights.}
\label{fig:msgm}
\vspace{-0.05in}
\end{figure}

\subsection{Multi-scale gated module}\label{sec:cornerpoint3d-multigated}

The MSGM is designed for adaptive feature extraction from LiDAR point clouds with different densities and properties. As shown in Figure~\ref{fig:msgm}, multi-scale 2D convolutions are utilized to extract features from the BEV features say $\mathcal{F}_{\text{BEV}}$ under three different receptive fields, allowing the model to adapt to point cloud inputs with varying sparsity levels. A gated weight generation module is then employed to calculate the weights for feature fusion. We utilize three kernel sizes for the multi-scale convolutions, \ie 1, 3, and 5, to effectively extract features from dense and sparse point cloud data. For the gated weight generation module, a global average pooling layer is first adopted, followed by two fully connected layers and a softmax function. The detailed process is shown below.

We first extract the multi-scale convolutional features say $\mathcal{F}_i$ from the BEV features $\mathcal{F}_{\text{BEV}}$ as
\begin{align}
\mathcal{F}_i = \text{Conv}_{i \times i}(\mathcal{F}_{\text{BEV}}),
\end{align}
where $i \in \{1, 3, 5\}$ represents different kernel sizes. Meanwhile, the BEV features $\mathcal{F}_{\text{BEV}}$ are fed into the gated weight generation module to obtain the gated weights, $\mathit{w}_i, i=1, 2, 3$, for the feature fusion of the three scales, \ie
\begin{align}
{w}_i =  \text{softmax} & \Big({W}_2 \cdot \text{ReLU}\big({W}_1 \cdot \text{GAP}(\mathcal{F}_{\text{BEV}}) \nonumber \\
&    + b_1 \big) + b_2\Big), \quad  i = 1, 2, 3,
\end{align}
where ReLU is the activation function between the two fully connected layers respectively with weights $W_1$ and $W_2$, GAP is the global average pooling, $b_1$ and $b_2$ are the biases. Finally, we obtain the multi-scale gated feature $\mathcal{F}_{\text{out}}$ by
\begin{align}
\mathcal{F}_{\text{out}} = \sum_{i \in \{1, 3, 5\}} w_i \cdot \mathcal{F}_i.
\end{align}
\subsection{Nearest corners heatmap generation} \label{subsec:nchg}

\begin{figure*}[t]
\begin{center}
   \includegraphics[width=0.85\linewidth]{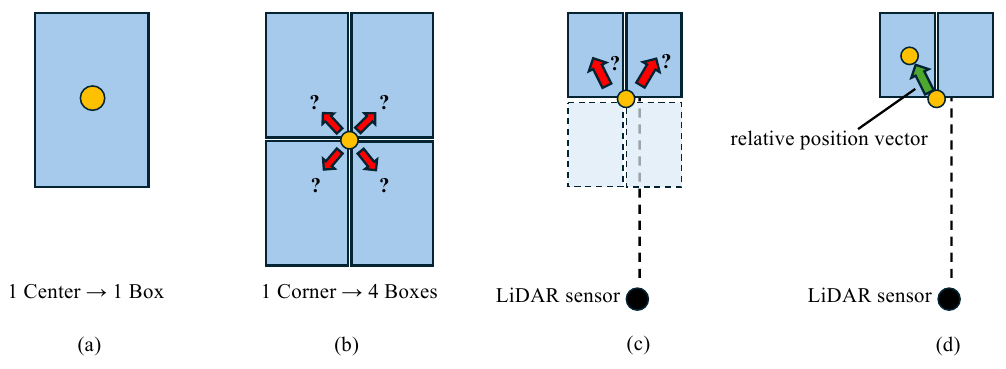}
   \vspace{-0.05in}
\end{center}
   \caption[Different conditions of the corner-box relationship]{Different conditions of the corner-box relationship. (a) Given a center point, it will only belong to one box. (b) Given a corner point, it may belong to four boxes. (c) By limiting the boxes to those where the point is the nearest one among the four corners, two box candidates still exist in some cases. (d) We utilize an additional separate head to predict the relative position vector between the nearest corner and the center of the same box, \ie extra guidance is provided to ensure accurate box selection.}
\label{fig:corner-to-boxes-note}
\end{figure*}

Following similar ideas to CenterPoint~\cite{Yin_2021_CVPR}, we generate a heatmap peak as the initial representative of detected objects in the BEV feature map from MSGM. In most cases, LiDAR sensors mainly capture the points that are not occluded by other objects or surfaces, which often surround the closer surfaces of objects. Therefore, predicting the nearest corner is more reasonable than predicting the center of objects, as it makes better use of the information contained in the point cloud data. With this in mind, unlike CenterPoint which generates heatmaps for the center of objects, CornerPoint3D uses heatmaps to represent the objects’ nearest corners to the ego vehicle and the LiDAR sensor. We first calculate the distance between four corners of the object's ground-truth bounding box in the BEV plane and find the nearest one to the ego vehicle (\ie the coordinate origin). Afterwards, the ground-truth heatmaps are generated by applying a Gaussian kernel function to the locations of the nearest corners for each object class. Following CenterPoint~\cite{Yin_2021_CVPR}, we calculate the radius of the Gaussian kernel by defining the $\sigma$ in the function as 
\begin{equation}
\sigma = \max(f(wl), \tau),
\end{equation}
where the function $f$ is defined in CornerNet~\cite{Law_2018_ECCV} (see more details in Appendix~\ref{sec:preliminaties_cornerpoint3d}), and $\tau = 2$ is the smallest allowable Gaussian radius defined following CenterPoint. A focal loss~\cite{lin2017focal} is then used to supervise the quality of the heatmaps.

\subsection{Additional separate head}\label{sec:cornerpoint3d-separatehead}

As shown in Figure~\ref{fig:modelstructure}, once getting the heatmaps of the objects' nearest corners, we use six independent separate heads to regress the other core features of object bounding boxes. Unlike the center-based process in CenterPoint which relies on a one-to-one correspondence between a center and a bounding box, corner-based box generation faces a one-to-many issue. As illustrated in Figure~\ref{fig:corner-to-boxes-note} (a) and (b), given a corner point, there are four possible boxes it may belong to, which is more complex than the situation for a center point. Even if we limit the boxes to those where the corner point is the nearest one to the origin among four box corners, see Figure~\ref{fig:corner-to-boxes-note} (c), there are still two box candidates that match the requirements. To counteract this problem, we use an additional separate head to predict the relative position vector between the nearest corner and the center of the bounding boxes, as shown in Figure~\ref{fig:corner-to-boxes-note} (d), providing extra guidance on the selection of the bounding box according to the corners. The other five separate heads play the same role as those in CenterPoint, regressing the location refinement that reduces the error caused by discrete corner heatmaps, the height of the nearest corner, the sizes, and the rotation angle of the bounding boxes. 


\subsection{CornerPoint3D equipped with EdgeHead}

To further guide the models to focus more on the closer surfaces of objects where more points surround, we utilize the EdgeHead~\citep{zhang2024detect} as the second-stage refinement head of CornerPoint3D. As a refinement head, the EdgeHead can be equipped with most 3D detectors, including our CornerPoint3D. In detail, we feed the predictions into the EdgeHead as RoIs, while also forwarding the earlier backbone features extracted by CornerPoint3D. These backbone features are further aggregated into the RoI features in the EdgeHead, enriching the model’s understanding of spatial and contextual relationships within the 3D scene. We name this two-stage CornerPoint3D (with EdgeHead) as the CornerPoint3D-Edge.

\section{Experiments}

\begin{figure*}[t]
\centering
\includegraphics[width=\textwidth]{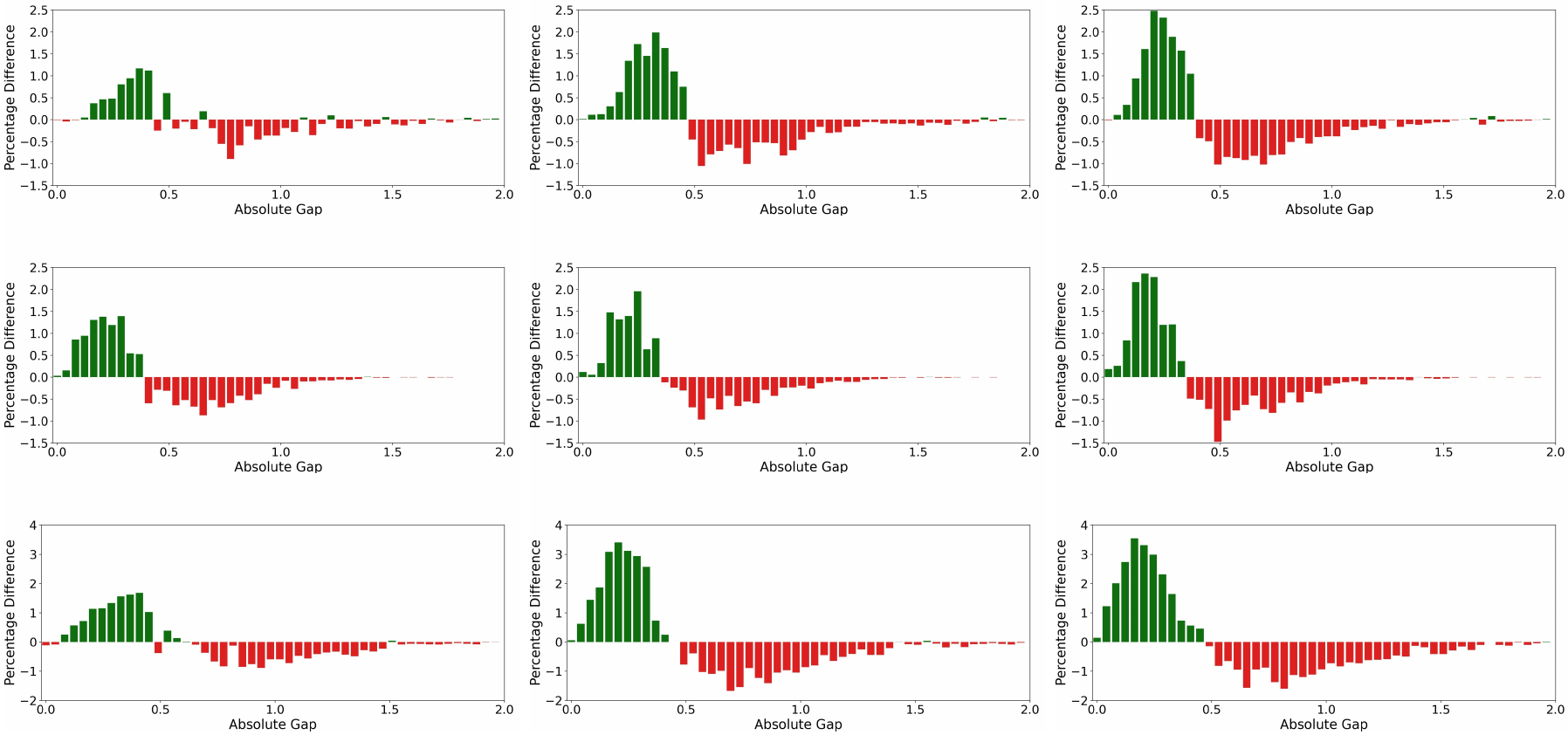}
\put(-460,160){\rotatebox{90}{\tiny Waymo $\rightarrow$ KITTI}}
\put(-460,85){\rotatebox{90}{\tiny Waymo $\rightarrow$ nuScenes}}
\put(-460,7){\rotatebox{90}{\tiny nuScenes $\rightarrow$ KITTI}}
\put(-433,142){\footnotesize (a) SECOND w/ EdgeHead}
\put(-288,142){\footnotesize (b) SECOND+ROS w/ EdgeHead}
\put(-130,142){\footnotesize (c) SECOND+SN w/ EdgeHead}
\put(-433,68){\footnotesize (d) CenterPoint w/ EdgeHead}
\put(-288,68){\footnotesize (e) CenterPoint+ROS w/ EdgeHead}
\put(-130,68){\footnotesize (f) CenterPoint+SN w/ EdgeHead}
\put(-435,-7){\footnotesize (g) SECOND w/ PointEdgeHead}
\put(-290,-7){\footnotesize (h) SECOND+ROS w/ PointEdgeHead}
\put(-133,-7){\footnotesize (i) SECOND+SN w/ PointEdgeHead}
\vspace{0.05in}
\caption{Proportion difference of the absolute gap of the closer surfaces, before and after equipped with (w/) EdgeHead. (a)--(c): SECOND in the Waymo $\rightarrow$ KITTI task. (d)--(f): CenterPoint in the Waymo $\rightarrow$ nuScenes task. (g)--(i): SECOND with our PointEdgeHead in the nuScenes $\rightarrow$ KITTI task. Columns one to three show the comparisons of models without domain adaptation/generalization methods, with the ROS, and with the SN, respectively. }
\label{fig:Gcs}
\vspace{-0.05in}
\end{figure*}
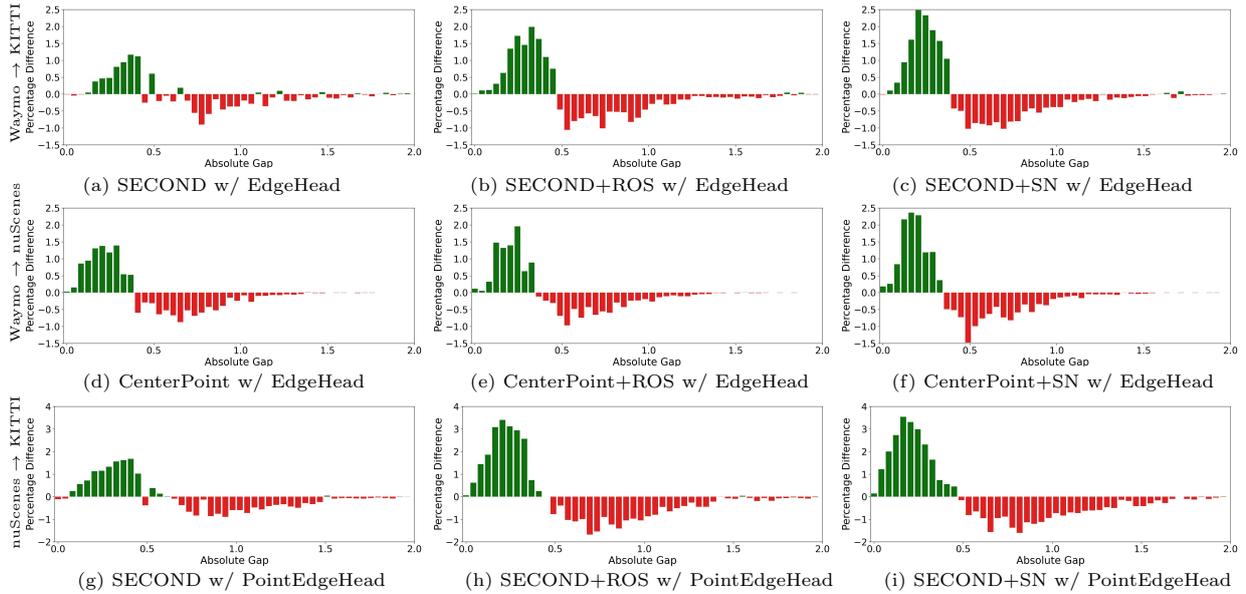

\subsection{Experimental setup} \label{sec:experiment_setup}

\noindent\textbf{Datasets.}
We conduct our main experiments on three datasets that have been widely used in 3D object detection tasks: KITTI~\cite{6248074}, nuScenes~\cite{9156412}, and Waymo~\cite{9156973}. As mentioned in ST3D~\cite{9578132}, KITTI only provides annotations in the front view, which makes it much more difficult to adapt models from KITTI to the other two datasets that provide ring view point cloud data and annotations. We therefore only take KITTI as the target domain and conduct our experiments in the following cross-domain tasks: nuScenes $\rightarrow$ KITTI, Waymo $\rightarrow$ KITTI and Waymo $\rightarrow$ nuScenes. The evaluation is focused on the \textit{Car} category (\ie the \textit{Vehicle} in Waymo), which has the most samples in all the datasets and has been the main focus in existing works~\cite{wang2020train, 9578132, zhang2024detect, wei2022lidar}.

\noindent\textbf{Data integration.} 
Harmonizing the differences between datasets is essential for cross-domain tasks. Key differences between datasets significantly influence cross-domain experiments, including (i) the point cloud range; (ii) the origin of coordinates; and (iii) the preprocessing units for point cloud data, such as voxel sizes in voxel-based methods. To address these issues, we adopt preprocessing techniques inspired by prior works~\cite{wang2020train, 9578132, zhang2024detect}. Specifically, we set the point cloud range for all datasets to $[-75.2, -75.2, -2.0, 75.2, 75.2, 4.0]$ meters and vertically shift the point cloud space of each dataset to ensure the X-Y plane aligns with the horizontal plane. Additionally, following ST3D~\cite{9578132} and EdgeHead~\cite{zhang2024detect}, we standardize the voxel size for voxel-based methods to $(0.1, 0.1, 0.15)$ meters across all datasets.

\noindent\textbf{Experiments for EdgeHead.} To evaluate the effect of our EdgeHead on cross-domain performance, we select two representative detectors to be equipped with the EdgeHead: (i) CenterPoint~\cite{Yin_2021_CVPR}, the well-known anchor-free model that CornerPoint3D is built upon; and (ii) SECOND~\cite{s18103337}, a representative anchor-based 3D object detection model that has inspired many voxel-based detectors. We first train these models as baselines on KITTI, Waymo, and nuScenes using the OpenPCDet~\citep{openpcdet2020} toolbox with suggested numbers of epochs and learning rates. In detail, we train them on KITTI for 80 epochs with learning rate $1\times10^{-3}$ and batch size of 8. With the learning rate unchanged, epochs of 50 and batch size of 16 are used for the training of the same models on nuScenes, and epochs of 30 and batch size of 8 for Waymo. For the second-stage refinement with our EdgeHead, we continue the training based on the pre-trained models for the same epochs as above, during which the parameters of the original models are frozen and only the EdgeHead is being trained. We also train the above models with two data augmentation methods -- ROS~\citep{9578132} and SN~\citep{9156543} -- and combine them with our EdgeHead. We use the same training settings as their original reproductions on OpenPCDet. Following other works~\citep{ bai2021pointdsc, 9578132} based on OpenPCDet, we adopt random horizontal flip, rotation, and scale transforms during the training process. All experiments are conducted on RTX 8000.

\noindent\textbf{Experiments for CornerPoint3D.} 
We train our proposed CornerPoint3D with the same epochs and learning rates as for CenterPoint, as described above. We compare the performance of CornerPoint3D with the same two methods as for EdgeHead, \ie the SECOND~\citep{s18103337} and CenterPoint~\citep{Yin_2021_CVPR}. In the main paper, we mainly discuss the comparison with CenterPoint. The comparison with SECOND is reported in the Appendix.

\noindent\textbf{Evaluation metrics.} Our evaluation includes two parts. Firstly, we evaluate the models' performance under the commonly used BEV and 3D metrics, measuring the models' ability to detect the entire 3D bounding boxes. Secondly, we evaluate the models using our proposed two additional evaluation metrics, \ie the \APnew and \APcs, to measure the models' ability to detect the nearest corner of an object and its two surfaces that are closer to the LiDAR sensor. 

\begin{table*}[htbp]
    \caption{Main comparisons for SECOND and CenterPoint across different tasks. We report \APBEV,  \APthreeD and \APcs of the car category at IoU = 0.7 and \APnew at IoU = 0.5. The reported performance is the moderate case when KITTI is the target domain, and is the overall result for other cross-domain tasks. Improvement (\ie fifth column) is calculated by the relative difference between each used method and the original model (\ie the first row of each task). }
    \label{tab:main_comparison}
    \vspace{-0.05in}
    \setlength\tabcolsep{5pt}
    \centering
    \resizebox{0.99\textwidth}{!}{
        \begin{tabular}{c c | c c c | c c c}
            \toprule[1pt]
            \multirow{3}{*}{Task} & \multirow{3}{*}{Method}  & \multicolumn{3}{c}{SECOND} & \multicolumn{3}{c}{CenterPoint} \\
            &  & \APBEV / \APthreeD & \APnew / \APcs & Improvement (\%) & \APBEV / \APthreeD & \APnew / \APcs & Improvement (\%) \\
            \midrule
            \multirow{6}{*}{W $\rightarrow$ K} 
            & Original  & 49.2 / 9.3 & 19.0 / 10.9 & - & 51.3 / 13.1 & 18.2 / 9.5 & - \\
            & + EdgeHead & 52.3 / 10.7 & 23.7 / 14.7 & 24.7\% / 34.9\% & 53.9 / 14.5 & 22.0 / 13.3 & 20.9\% / 40.0\% \\
            & + ROS & 73.0 / 38.3 & 33.7 / 12.6 & 77.4\% / 15.6\% & 75.1 / 44.2 & 41.1 / 19.1 & 126.1\% / 101.1\% \\
            & + EdgeHead \& ROS & 76.4 / 41.5 & 42.9 / 20.4 & 125.8\% / 87.2\% & 77.3 / 47.4 & 46.2 / 23.2 & 154.4\% / 144.2\% \\
            & + SN & 73.0 / 55.5 & 49.3 / 20.5 & 159.5\% / 87.9\% & 72.5 / 56.7 & 51.4 / 24.9 & 182.4\% / 162.1\% \\
            & + EdgeHead \& SN & 79.7 / 64.2 & 62.3 / 34.2 & 227.9\% / 213.8\% & 77.8 / 63.5 & 59.8 / 30.1 & 228.6\% / 216.8\% \\
            \midrule
            \multirow{6}{*}{W $\rightarrow$ N} 
            & Original  & 27.8 / 16.1 & 15.6 / 6.7 & - & 30.4 / 16.7 & 19.9 / 12.3 & - \\
            & + EdgeHead & 29.9 / 18.0 & 20.9 / 13.0 & 34.0\% / 94.0\% & 29.7 / 17.6 & 21.3 / 13.8 & 7.0\% / 12.2\% \\
            & + ROS & 26.7 / 15.4 & 15.8 / 6.5 & 1.3\% / -3.0\% & 28.8 / 16.2 & 19.5 / 11.7 & -2.0\% / -4.9\% \\
            & + EdgeHead \& ROS & 28.3 / 17.1 & 19.9 / 11.9 & 27.6\% / 77.6\% & 29.2 / 17.4 & 21.3 / 13.4 & 7.0\% / 9.3\% \\
            & + SN & 26.4 / 16.4 & 16.7 / 8.7 & 7.1\% / 29.9\% & 29.4 / 18.0 & 20.5 / 12.7 & 3.0\% / 3.3\% \\
            & + EdgeHead \& SN & 28.4 / 18.6 & 20.7 / 13.4 & 32.7\% / 100.0\% & 29.3 / 19.2 & 22.1 / 18.9 & 11.1\% / 53.7\% \\
            \midrule
            \multirow{6}{*}{N $\rightarrow$ K} 
            & Original  & 35.7 / 11.8 & 16.4 / 9.8 & - & 34.6 / 8.3 & 13.1 / 5.8 & - \\
            & + EdgeHead & 53.6 / 15.9 & 33.3 / 19.6 & 103.0\% / 100.0\% & 37.0 / 10.4 & 19.6 / 11.5 & 49.6\% / 98.3\% \\
            & + ROS & 43.4 / 20.0 &  20.2 / 8.1 & 23.2\% / -17.3\% & 43.8 / 20.6 & 27.6 / 13.1 & 110.8\% / 125.9\% \\
            & + EdgeHead \& ROS & 52.7 / 33.1 & 39.9 / 24.6 & 143.3\% / 151.0\% & 60.3 / 31.3 & 43.2 / 21.3 & 229.8\% / 267.2\% \\
            & + SN & 29.6 / 14.3 & 15.7 / 8.2 & -4.3\% / -16.3\% & 33.5 / 18.1 & 22.0 / 11.6 & 67.9\% / 100.0\% \\
            & + EdgeHead \& SN & 45.7 / 30.4 & 35.1 / 23.5 & 114.0\% / 139.8\% & 58.4 / 34.7 & 44.8 / 26.8 & 241.2\% / 362.1\% \\
            \bottomrule[0.8pt]
        \end{tabular}
       } 
\end{table*}

\begin{table*}[t]
    \caption[Main results of CornerPoint3D and CornerPoint3D-Edge, and comparisons with CenterPoint under different tasks]{Main results of CornerPoint3D and CornerPoint3D-Edge, and comparisons with CenterPoint~\cite{Yin_2021_CVPR} under different tasks. We report \APBEV,  \APthreeD and \APcs of the car category at IoU = 0.7 and \APnew at IoU = 0.5. The reported performance is the moderate case when KITTI is the target domain, and is the overall result for other cross-domain tasks. Improvements are calculated by the relative difference between each CenterPoint and CornerPoint3D equipped with the same methods (\eg native, the EdgeHead (Edge)~\cite{zhang2024detect}, and the ROS~\cite{wang2020train}).  W, K, and N represent the Waymo, KITTI, and nuScenes datasets, respectively. }
    \label{tab:cornerpoint3d_with_edgehead}
    \small
    \setlength\tabcolsep{6pt}
    \centering
    \resizebox{0.96\textwidth}{!}{
    \begin{tabular}{l l l l l l}
    \toprule[1pt]
    Task & Method & \APBEV & \APthreeD & \APnew & \APcs \\
    \midrule
    \multirow{8}{*}{W $\rightarrow$ K} 
    & CenterPoint & 51.3 & 13.1 & 18.2 & 9.5  \\
    & \textbf{CornerPoint3D} & \textbf{47.5} (-7.4\%) & \textbf{8.4} (-35.9\%) & \textbf{20.0} (+9.9\%) & \textbf{11.6} (+22.1\%) \\
    & CenterPoint w/ Edge & 53.9 & 14.5 & 22.0 & 13.3 \\
    & \textbf{CornerPoint3D-Edge} & \textbf{58.9} (+9.3\%) & \textbf{12.4} (-14.5\%) & \textbf{28.3} (+28.6\%) & \textbf{18.6} (+39.8\%) \\
    & CenterPoint w/ ROS & 75.1 & 44.2 & 41.1 & 19.1 \\
    & \textbf{CornerPoint3D w/ ROS} & \textbf{70.0} (-6.8\%) & \textbf{29.9} (-32.4\%) & \textbf{39.2} (-4.6\%) & \textbf{19.3} (+1.0\%) \\
    & CenterPoint w/ Edge \& ROS & 77.3 & 47.4 & 46.2 & 23.2 \\
    & \textbf{CornerPoint3D-Edge w/ ROS} & \textbf{80.7} (+4.4\%) & \textbf{48.7} (+2.7\%) & \textbf{50.9} (+10.2\%) & \textbf{24.8} (+6.9\%) \\
    \midrule
    \multirow{8}{*}{W $\rightarrow$ N} 
    & CenterPoint & 30.4 & 16.7 & 19.9 & 12.3  \\
    & \textbf{CornerPoint3D} & \textbf{25.8} (-15.1\%) & \textbf{11.6} (-30.5\%) & \textbf{15.9} (-20.1\%) & \textbf{9.1} (-26.0\%) \\
    & CenterPoint w/ Edge & 29.7 & 17.6 & 21.3 & 13.8 \\
    & \textbf{CornerPoint3D-Edge} & \textbf{29.8} (+0.3\%) & \textbf{15.0} (-14.8\%) & \textbf{21.2} (-0.5\%) & \textbf{13.0} (-5.8\%) \\
    & CenterPoint w/ ROS & 28.8 & 16.2 & 19.5 & 11.7 \\
    & \textbf{CornerPoint3D w/ ROS} & \textbf{24.7} (-14.2\%) & \textbf{10.2} (-37.0\%) & \textbf{14.5} (-25.6\%) & \textbf{7.7} (-34.2\%) \\
    & CenterPoint w/ Edge \& ROS & 29.2 & 17.4 & 21.3 & 13.4 \\
    & \textbf{CornerPoint3D-Edge w/ ROS} & \textbf{29.7} (+1.7\%) & \textbf{15.3} (-12.1\%) & \textbf{21.2} (-0.5\%) & \textbf{13.3} (-0.8\%) \\
    \midrule
    \multirow{8}{*}{N $\rightarrow$ K} 
    & CenterPoint & 34.6 & 8.3 & 13.1 & 5.8  \\
    & \textbf{CornerPoint3D} & \textbf{43.0} (+24.6\%) & \textbf{12.5} (+50.6\%) & \textbf{24.1} (+84.0\%) & \textbf{12.9} (+122.4\%) \\
    & CenterPoint w/ Edge & 37.0 & 10.4 & 19.6 & 11.5 \\
    & \textbf{CornerPoint3D-Edge} & \textbf{49.7} (+34.3\%) & \textbf{18.5} (+77.9\%) & \textbf{32.5} (+65.8\%) & \textbf{21.9} (+90.4\%) \\
    & CenterPoint w/ ROS & 43.8 & 20.6 & 27.6 & 13.1 \\
    & \textbf{CornerPoint3D w/ ROS} & \textbf{42.6} (-2.7\%) & \textbf{20.1} (-2.4\%) & \textbf{30.9} (+12.0\%) & \textbf{19.2} (+46.6\%) \\
    & CenterPoint w/ Edge \& ROS & 60.3 & 31.3 & 43.2 & 21.3 \\
    & \textbf{CornerPoint3D-Edge w/ ROS} & \textbf{65.3} (+8.3\%) & \textbf{41.9} (+33.9\%) & \textbf{53.2} (+23.2\%) & \textbf{35.4} (+66.2\%) \\
    \bottomrule[0.8pt]
\end{tabular}
       }
\end{table*}

\subsection{Main results of EdgeHead}

In this section, we demonstrate the main results of our proposed EdgeHead. We first compare the proposed EdgeHead with two representative detectors, the CenterPoint~\citep{Yin_2021_CVPR} and SECOND~\citep{s18103337}, by measuring their absolute gaps of the closer surfaces (\ie $G_{\rm cs}$). Afterwards, we analyze the performance of these two detectors before and after integrating EdgeHead.

\subsubsection{The absolute gap of the closer surfaces}

In this section, we compare the proposed EdgeHead with existing methods by measuring their absolute gaps of the closer surfaces (\ie $G_{\rm cs}$). As shown in Figure~\ref{fig:Gcs}, we calculate the distributions of $G_{cs}$ for each comparison pair of methods and draw the proportion difference between them. Specifically, we quantify the $G_{\rm cs}$ distribution of two models within an identical interval $I$ (set to $[0,2]$ by default), and then calculate the proportion difference as
$
    {\rm Diff}_{AB}(i) = P_B^i - P_A^i,
$
where $P_{A}^i$ and $P_{B}^i$ denote the proportion of $G_{\rm cs}$ in the $i\text{-th}$ sub-interval of $I$ for models $A$ and $B$, respectively. Therefore, if the left part of the proportion difference graph is above the X-axis and the right half is vice versa, we can tell that model B predicts the closer surfaces better than model A and thus has a $G_{\rm cs}$ distribution closer to zero. For example, Figure~\ref{fig:Gcs}(a) shows that SECOND+EdgeHead (\ie the SECOND model combined with our proposed EdgeHead) predicts the closer surfaces better than the original SECOND model when trained on Waymo and tested on KITTI. Consistent results are observed for the other tasks and models, which demonstrates that our EdgeHead can stably shift the $G_{cs}$ distribution to the left, \ie improve the detection quality regarding the closer surfaces. 

We also plot the proportion difference to analyze models using data augmentation methods (\ie ROS and SN; see the last two columns in Figure~\ref{fig:Gcs}), which will be further discussed in Sections~\ref{sec:rossn}.


\subsubsection{EdgeHead equipped with existing models}

Table~\ref{tab:main_comparison} presents the quantitative comparison of the performance between different models before and after equipping with our proposed EdgeHead. It shows that models equipped with EdgeHead can achieve better \APcs and \APnew than the original models in all cross-domain tasks. As described in Section~\ref{sec:metricdesign}, the \APcs directly measures the improvement in the detection quality of the closer surfaces, and the consistently improved performance under this metric shows that EdgeHead can stably improve the closer-surfaces detection ability of existing models across various domains. {The results of the \APcs are also consistent with the proportion difference of the closer-surfaces absolute gap shown in Figure~\ref{fig:Gcs}.} Meanwhile, the improvement under the \APnew metric shows that EdgeHead also works well when evaluating models with a balance between the accuracy of the entire box and the closer surfaces.

Table~\ref{tab:main_comparison} also shows that the models' performance changes much less or even remains at the original value level when evaluated under the original BEV and 3D metrics. Taking the SECOND model in the Waymo $\rightarrow$ KITTI task as an example, the BEV AP and 3D AP respectively improved by 6.3\% (from 49.2 to 52.3) and 15.1\% (from 9.3 to 10.7) when equipped with EdgeHead, while the \APcs and \APnew represent the improvement by 24.7\% and 34.9\%, respectively. We also summarize the gaps between each original model before and after equipping with EdgeHead in Table~\ref{tab:main_comparison}, which shows that similar phenomena can also be observed in other comparisons. The larger improvement shown in the \APcs and \APnew supports two important conclusions: (i) the newly proposed metrics are truly more sensitive to the closer-surfaces detection ability, and therefore can evaluate the model's cross-domain performance from a different point of view; and (ii) our EdgeHead can effectively improve the model's ability to detect the closer surfaces, which is truly helpful for applications in cross-domain tasks.

\subsection{CornerPoint3D}

In this section, we extensively evaluate our detector CornerPoint3D on different cross-domain tasks, under the four different metrics. Similar to last section, we compare the performance of CornerPoint3D with the representative 3D detectors, CenterPoint~\cite{Yin_2021_CVPR}, and SECOND~\cite{s18103337}. We illustrate the main results in Table~\ref{tab:cornerpoint3d_with_edgehead}. 

For the Waymo $\rightarrow$ KITTI task where existing a big gap in the average car sizes, CornerPoint3D realizes a trade-off between the detection quality of entire bounding boxes (\ie measured by the \APBEV and \APthreeD) and the locating accuracy of closer surfaces to the ego vehicle (\ie measured by the \APnew and \APcs). By focusing more on the closer surfaces via the corner-based heatmaps, higher \APnew and \APcs are achieved with an acceptable drop in \APBEV and \APthreeD, resulting in a similar detection quality of entire boxes to the SECOND detector. For the Waymo $\rightarrow$ nuScenes task, \ie from a denser domain to a sparse domain, our CornerPoint3D does not outperform the CenterPoint detector, while achieving comparable performance with SECOND (a lower \APthreeD and a higher \APcs). As already discussed in existing works~\cite{9578132, wei2022lidar}, it is hard to adapt detectors from the point clouds that are denser or with more LiDAR beams (\eg Waymo) to the point clouds that are sparser or with fewer LiDAR beams (\eg nuScenes), while the opposite adaptation is relatively much easier. In the nuScenes $\rightarrow$ KITTI task, CornerPoint3D achieves impressive performance in all four evaluation metrics, improving the \APnew and \APcs by 47.0\% and 31.6\%, respectively. It is worth noting that the Waymo and nuScenes datasets share more similar size distribution properties, which differ significantly from those in the KITTI dataset. Our experimental results on Waymo $\rightarrow$ KITTI and nuScenes $\rightarrow$ KITTI demonstrate that CornerPoint3D exhibits strong adaptability across domains with significant object size disparities, even when there are substantial density differences, provided the adaptation is from a sparser domain to a denser one.

\subsection{CornerPoint3D equipped with EdgeHead} 

In this section, we evaluate the cross-domain performance of CornerPoint3D-Edge, \ie the two-stage CornerPoint3D with EdgeHead being equipped as the second-stage refinement head, with the main results in Table~\ref{tab:cornerpoint3d_with_edgehead}.

Applying the EdgeHead~\cite{zhang2024detect} to CornerPoint3D improves the performance much more (\ie by approximately 20\%) compared with applying it to CenterPoint across all three cross-domain tasks. For example, in the Waymo $\rightarrow$ KITTI task, CornerPoint3D achieves a score of 20.0 under \APnew, which is 9.9\% higher than CenterPoint. After integrating the EdgeHead, CornerPoint3D-Edge achieves 28.3 under \APnew, widening the gap to 28.6\% compared to CenterPoint with the EdgeHead. Similar performance improvements are observed across all three cross-domain tasks and under all four evaluation metrics, demonstrating the enhanced synergy between CornerPoint3D and EdgeHead.

\subsection{Compatibility of EdgeHead}\label{sec:rossn}

ROS~\cite{9578132} and  SN~\cite{9156543} are two data augmentation methods in cross-domain tasks that aim to solve the overfitting problems in object sizes. ROS randomly scales the size of object boxes in both the annotations and the point cloud data to make the model more robust to object sizes. SN uses the average object size of each dataset as additional information and normalizes the source domain's object size by using the target domain's size statistics. It is therefore worth investigating the influence of such methods on the models' closer-surfaces detection ability, and their compatibility with our EdgeHead. 

We below first evaluate the CS-ABS AP and CS-BEV AP of different models equipped with these two methods (\ie ROS and SN) including further combining them with our EdgeHead. We denote these combinations as +ROS, +SN, +EdgeHead \& ROS and +EdgeHead \& SN in Table~\ref{tab:main_comparison}. The comparisons of the absolute gap are shown in Figure~\ref{fig:Gcs} as well. For most tasks, ROS and SN can help the models achieve higher BEV AP and 3D AP, but cannot stably improve the CS-ABS AP and CS-BEV AP by a similar margin. Taking SECOND in the Waymo $\rightarrow$ KITTI task as an example, ROS increases the BEV AP and 3D AP respectively by 48.4\% and 311.8\% (\ie from 49.2 / 9.3 to 73.0 / 38.3) but only increases the CS-ABS AP by 15.6\%. In comparison, the additional use of our EdgeHead increases the performance under all four metrics, especially for the CS-ABS AP and CS-BEV AP. Taking the above example, the performance under the new metrics increases by 125.8\% and 87.2\% when equipping SECOND with ROS and EdgeHead simultaneously. Consistent results can also be observed for the other tasks and models in Table~\ref{tab:main_comparison} and Figure~\ref{fig:Gcs}.
%
We also noticed that for both models in the Waymo $\rightarrow$ nuScenes task, ROS and SN increase the performance much less or even decrease it due to the minor object size difference between these two datasets, which is also mentioned in \cite{9578132}. However, the performance can still be greatly improved by using our EdgeHead and ROS / SN together.

The above results demonstrate that our proposed EdgeHead can be effectively used with the existing data augmentation methods designed for the size overfitting problem, which not only further improves the models' detection ability for the entire box but also helps achieve much better closer-surfaces detection quality compared with only using these data augmentation methods.

\subsection{Compatibility of CornerPoint3D}

In this section, we investigate the effect of applying the data augmentation method ROS~\cite{wang2020train} to our CornerPoint3D and CornerPoint3D-Edge, as an example. As previously, we take CenterPoint~\cite{Yin_2021_CVPR} as the control group for comparison. As shown in Table~\ref{tab:cornerpoint3d_with_edgehead}, we first apply the ROS to CenterPoint and CornerPoint3D (\ie w/ ROS), and then apply the EdgeHead and the ROS to the CenterPoint at the same time (\ie w/ EdgeHead \& ROS), to compare with the CornerPoint3D-Edge equipped with ROS. 


\noindent\textbf{ROS with CornerPoint3D.} Applying the ROS to CornerPoint3D leads to a great performance increase in both the Waymo $\rightarrow$ KITTI and the nuScenes $\rightarrow$ KITTI tasks, although it actually has a better influence when applied to CenterPoint. Specifically, after incorporating ROS, CornerPoint3D achieves a score of 39.2 under \APnew in the Waymo $\rightarrow$ KITTI task, which is even higher than when equipped with the EdgeHead (28.3). However, applying ROS to CenterPoint improves its \APnew from 22.0 to 41.1, resulting in a greater performance gain. Additionally, the performance improvement is not as stable as brought by the EdgeHead. For example, in the nuScenes $\rightarrow$ KITTI task, applying ROS leads to a lower performance increase (\eg 24.1 $\rightarrow$ 30.9 under \APnew) than applying the EdgeHead (\eg 24.1 $\rightarrow$ 32.5 under \APnew); and in the Waymo $\rightarrow$ nuScenes task, it even leads to lower performance than the identical CornerPoint3D (\eg 15.9 $\rightarrow$ 14.5 under \APnew). Although ROS also aims to enhance model robustness and performance in cross-domain scenarios, it is designed for entire bounding box detection based on center points. As a result, it is less effective for CornerPoint3D, which relies on a nearest corner box representation and focuses more on the detection of closer object surfaces.

\noindent\textbf{ROS with CornerPoint3D-Edge.} Encouragingly, applying ROS to our CornerPoint3D-Edge proves to be the most stable and effective strategy. When ROS is applied to CornerPoint3D-Edge, it achieves the highest performance in the Waymo $\rightarrow$ KITTI and nuScenes $\rightarrow$ KITTI tasks, surpassing CenterPoint with both ROS and EdgeHead across all four evaluation metrics. For instance, CornerPoint3D-Edge with ROS achieves \APnew scores of 50.9 and 53.2 in these two tasks, respectively, representing improvements of 10.2\% and 23.2\ over CenterPoint equipped with the same methods (\ie 46.2 and 43.2).

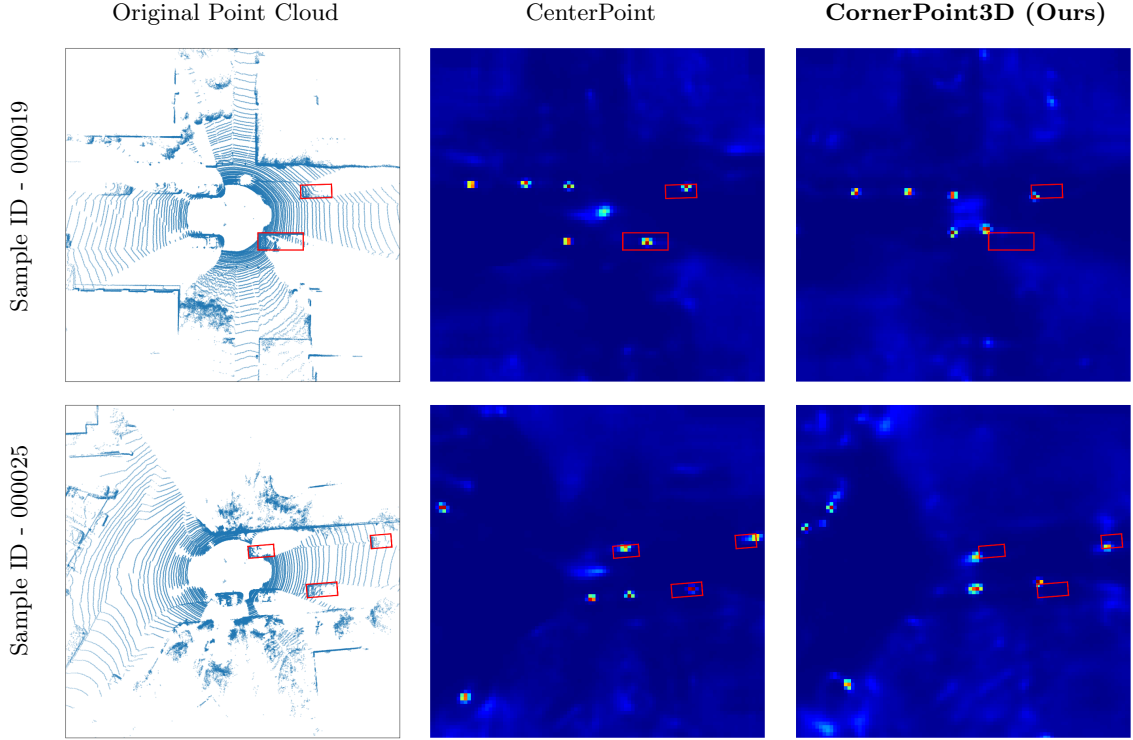
\begin{figure*}[t]
\begin{center}
   \includegraphics[width=0.9\linewidth]{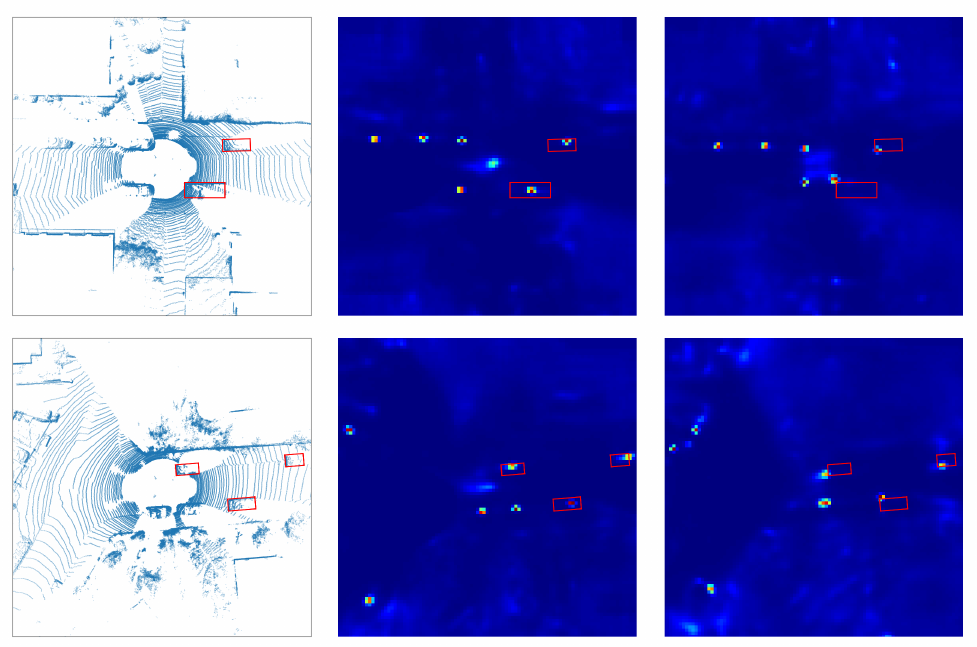}
   \put(-387,275){{\small Original Point Cloud}}
   \put(-232,275){{\small CenterPoint}}
   \put(-120,275){{\small \textbf{CornerPoint3D (Ours)}}}
   \put(-425,165){\rotatebox{90}{\small Sample ID - 000019}}
   \put(-425,35){\rotatebox{90}{\small Sample ID - 000025}}
   \vspace{-0.05in}
\end{center}
   \caption[Nearest corner heatmaps and center heatmaps from CornerPoint3D and CenterPoint respectively]{Comparison between the nearest corner heatmaps (right column) from CornerPoint3D (ours) and the center heatmaps (middle column) from CenterPoint~\cite{Yin_2021_CVPR}. Visualizations of the original point cloud data are also provided (left column), magnified by 400\%. Both models are for the Waymo $\rightarrow$ KITTI task (\ie trained on Waymo and tested on KITTI). More examples are attached in the Appendix.}
\label{fig:heatmaps_w2k}
\end{figure*}

\begin{table}[hbpt]
\caption[Within-domain Performance of CornerPoint3D, CenterPoint and SECOND]{Within-domain performance of CornerPoint3D (Ours), CenterPoint (Center), and SECOND under four metrics on Waymo (W), nuScenes (N), and KITTI (K). Performance of CenterPoint and CornerPoint3D with EdgeHead is also included as Center w/ Edge and Ours-Edge, respectively.}
\label{tab:within-domain}
\setlength\tabcolsep{1.5pt}
\footnotesize
\centering
\begin{tabular}{c c c c c c} 
\toprule
Task & Method & \APBEV & \APthreeD & \APnew & \APcs \\
\midrule
\multirow{5}{*}{W}  
& SECOND & 63.1 & 47.5 & 47.3 & 34.2 \\
& CenterPoint & 67.3 & 52.5 & 52.6 & 41.4 \\
& \textbf{Ours} & \textbf{63.2} & \textbf{46.5} & \textbf{49.0} & \textbf{37.9} \\
& Center w/ Edge & 65.4 & 53.8 & 56.0 & 46.7 \\
& \textbf{Ours-Edge} & \textbf{67.4} & \textbf{53.1} & \textbf{56.5} & \textbf{47.2} \\
\midrule
\multirow{5}{*}{N}    
& SECOND & 43.1 & 22.6 & 25.9 & 14.2 \\
& CenterPoint & 49.7 & 30.7 & 34.1 & 23.2 \\
& \textbf{Ours} & \textbf{45.3} & \textbf{26.8} & \textbf{31.0} & \textbf{21.1} \\
& Center w/ Edge & 48.1 & 32.5 & 36.9 & 26.9 \\
& \textbf{Ours-Edge} & \textbf{49.7} & \textbf{32.4} & \textbf{37.6} & \textbf{28.2} \\
\midrule
\multirow{5}{*}{K} 
& SECOND & 84.3 & 72.1 & 71.4 & 53.3 \\
& CenterPoint & 84.4 & 73.6 & 72.8 & 56.8 \\
& \textbf{Ours} & \textbf{84.5} & \textbf{73.1} & \textbf{72.8} & \textbf{59.8} \\
& Center w/ Edge & 84.4 & 75.3 & 74.7 & 61.8 \\
& \textbf{Ours-Edge} & \textbf{86.9} & \textbf{77.5} & \textbf{80.7} & \textbf{68.9} \\
\bottomrule
\end{tabular}
\end{table}

\subsection{Corner-based heatmap}

In this section, we deeply analyze the corner-based learning process of the CornerPoint3D by visualizing the nearest-corner heatmaps and comparing them with the center heatmaps from the CenterPoint~\cite{Yin_2021_CVPR}. As shown in Figure~\ref{fig:heatmaps_w2k}, we visualize the corner and center heatmaps of CornerPoint3D and CenterPoint, respectively, with the same data sample. For objects that are closer to the LiDAR sensor (\ie the ego vehicle), both the nearest corners and the centers are accurately predicted in the heatmaps. However, for objects with a farther distance to the LiDAR sensor, the nearest corners are predicted much more accurately than the centers. Due to the longer distance from the sensor, fewer LiDAR points are available around the objects, resulting in significantly higher uncertainty in size estimation and more severe overfitting in center predictions. In contrast, while farther objects have fewer points, most of them are still concentrated around the closer surfaces of objects, which helps preserve the prediction accuracy of the nearest corners in the CornerPoint3D. The comparison between the nearest corner and the center heatmaps further indicates the robustness of CornerPoint3D in cross-domain tasks.

\subsection{Results on within-domain tasks}

In this section, we additionally compare the within-domain performance of the CornerPoint3D and two representative 3D object detectors,  CenterPoint~\cite{Yin_2021_CVPR} and  SECOND~\cite{s18103337}. As illustrated in Table~\ref{tab:within-domain}, CornerPoint3D achieves better results than SECOND in all three within-domain tasks, especially under the \APnew and \APcs metrics. Although lower performance is observed when comparing CornerPoint3D with the CenterPoint, it achieves better performance when simultaneously equipped with the EdgeHead~\cite{zhang2024detect}, indicating better compatibility with the additional refinement head. Consistent performance in all three within-domain tasks indicates that the proposed CornerPoint3D is not only designed for cross-domain 3D object detection tasks, but also maintains stable and competitive detection ability in traditional within-domain tasks. With both methods focusing on the closer surfaces of objects, where more prior knowledge (\ie point cloud data) is available, the successful integration of CornerPoint3D and EdgeHead further validates the effectiveness of our main ideas about the closer surfaces and the nearest corners of objects.

\begin{table}[tbp]
\caption{Ablation study of our EdgeHead under the SECOND model. Refine: Using a second stage refinement head. Corner: Replacing the center locations with the locations of the closest vertex in the refinement head. We use the proposed \APnew and \APcs metrics.}
\setlength\tabcolsep{2pt}
\footnotesize
\label{tab:controlgroup}
\centering
\begin{tabular}{c c c c c c} 
\toprule
Method                         & Refine & Corner & W $\rightarrow$ K & W $\rightarrow$ N & N $\rightarrow$ K\\ 
\midrule
     Original  &  \XSolidBrush & \XSolidBrush & 19.0 / 10.9   & 15.6 / 6.7 & 16.4 / 9.8 \\

     Control-group  & \Checkmark & \XSolidBrush & 21.6 / 12.1   & 18.7 / 11.4 & 19.2 / 10.1 \\

     EdgeHead  & \Checkmark & \Checkmark &  23.7 / 14.7   & 20.9 / 13.0 & 33.3 / 19.6 \\
     
\bottomrule
\end{tabular}
\end{table}

\begin{table*}[htbp]
\caption{Results of SECOND before and after equipping with EdgeHead under the CS-ABS and CS-BEV metrics with different $\alpha$ settings. The results on the Waymo $\rightarrow$ KITTI task are reported in this table. }

\label{tab:appendix_alpha}
\centering
\footnotesize
\begin{tabular}{c c c c c c } 
\toprule
Tasks                         & Metrics & SECOND & SECOND + EdgeHead\\ 
            &   & \APnew / \APcs  & \APnew / \APcs \\
\midrule
\multirow{3}{*}{$\alpha = 0.5 $}  
     & Original  &  63.8 / 70.5   & 69.0 / 74.8 \\
     & + ROS     & 75.1 / 75.2  & 79.8 / 81.6  \\
     & + SN    &  72.8 / 72.2  & 80.0 / 80.0 \\ 
     
\midrule
\multirow{3}{*}{$\alpha = 1 $}    
    & Original  &     19.0 / 10.9   &  23.7 / 14.7  \\
    & + ROS  &     33.7 / 12.6     &  42.9 / 20.4 \\
    & + SN   & 49.3 / 20.5   &  62.3 / 34.2 \\ 
    
\midrule
\multirow{3}{*}{$\alpha = 1.5 $} 
    & Original  &  3.8 / 1.0 &   4.8 / 1.4    \\
    & + ROS  &  9.6 / 1.2 &  14.6 / 2.3    \\
    & + SN &   21.8 / 3.3  &   35.8 / 6.9  \\

\bottomrule
\end{tabular}
\centering
\end{table*}

\begin{table*}[htbp]
\caption[Ablation study of our CornerPoint3D in the nuScenes $\rightarrow$ KITTI task]{Ablation study of our CornerPoint3D in the nuScenes $\rightarrow$ KITTI task. CornerHM: Using the nearest corner heatmap instead of the center heatmap as in CenterPoint~\cite{Yin_2021_CVPR}. MSGM: Using the proposed multi-scale gated module.}
\setlength\tabcolsep{4pt}
\footnotesize
\label{tab:controlgroup_corner3d}
\centering
\resizebox{0.99\textwidth}{!}{
\begin{tabular}{c c c c c c c c} 
\toprule
Method                         & CornerHM & EdgeHead & MSGM & \APBEV & \APthreeD & \APnew & \APcs\\ 
\midrule
    CenterPoint  &  \XSolidBrush & \XSolidBrush & \XSolidBrush & 34.6 & 8.3 & 13.1 & 5.8 \\

     Control-group 1  &  \Checkmark & \XSolidBrush & \XSolidBrush & 35.8 (+3.5\%) & 6.0 (-27.7\%)  & 17.6 (+34.4\%) & 9.7 (+67.2\%) \\

     Control-group 2  &  \XSolidBrush & \Checkmark & \XSolidBrush & 37.0 (+6.9\%) & 10.4 (+25.3\%)  & 19.6 (+49.6\%) & 11.5 (+98.3\%) \\
     Control-group 3  & \XSolidBrush & \XSolidBrush & \Checkmark & 40.4 (+16.8\%) & 13.5 (+62.7\%)  & 9.7 (-26.0\%) & 4.0 (-31.0\%) \\

    \textbf{CornerPoint3D}  & \Checkmark & \XSolidBrush & \Checkmark & 43.0 (+24.6\%) & 12.5 (+50.6\%)  & 24.1 (+84.0\%) & 12.9 (+122.4\%) \\

     \textbf{CornerPoint3D-Edge}  & \Checkmark & \Checkmark & \Checkmark &  49.7 (+43.6\%) & 18.5 (+122.9\%)  & 32.5 (+148.1\%) & 21.9 (+277.6\%) \\
     
\bottomrule
\end{tabular}
}
\centering
\end{table*}

\subsection{Ablation study} \label{sec:ablation_corner3d}

\subsubsection{Ablation study for EdgeHead} 

To further analyze our EdgeHead's refinement performance, below we propose another control-group head for us to conduct the ablation study. Specifically, we maintain the module structure and the loss function design of EdgeHead, while replacing the closest vertex in EdgeHead with the center coordinates for the calculation of the location regression target. Therefore, the loss function of the control-group head reads
\begin{align}
    {\cal L}_{\rm reg}^{\prime \prime} = \sum\limits_{r \in \left\{ x_{\rm c}, y_{\rm c}, \theta \right\} } {\cal L}_{{\rm smooth}-\ell_1}(\widehat{\Delta r^{a}}, \Delta r^{a}).
\end{align}
In other words, the above control-group head is a simplified version of the typical refinement module as described in Eq.~\eqref{eq:origin_loss}, which only refines the (BEV) location $x$, $y$, and the rotation angle $\theta$. By comparing the performance of EdgeHead and this control-group head, we can better understand the contribution of modifying the training purpose to the closer surfaces. As shown in Table~\ref{tab:controlgroup}, although the performance of the control-group head is better than the original model that does not use any refinement head, there is a rather significant gap in terms of detection performance improvement when comparing to the excellent results of our EdgeHead. The results in Table~\ref{tab:controlgroup} indicate that the regression target in our EdgeHead truly helps models achieve better closer-surfaces detection ability.

Furthermore, we analyze the different settings of $\alpha$ in the proposed CS-ABS and CS-BEV metrics. We set the value of $\alpha$ as 1 by default in the main paper, while we investigate the performance influence of different settings of $\alpha$ here. As shown in Table~\ref{tab:appendix_alpha}, when setting $\alpha$ to smaller values (\eg 0.5), it makes the penalty to the original BEV IoU smaller and thus provides closer CS-BEV results for the original SECOND model before and after equipping with EdgeHead. Conversely, setting $\alpha$ to larger values (\eg 1.5) could diminish the models' detection performance. Therefore, a more reasonable choice of $\alpha$ should be better in the range of $(0.5, 1.5)$ for practical use.

\subsubsection{Ablation study for CornerPoint3D} 

We finally conduct extensive ablation experiments to investigate the individual components of our CornerPoint3D. All experiments are conducted on the cross-domain task of nuScenes $\rightarrow$ KITTI. We equip the CenterPoint model with the nearest corner heatmap generation module (see Section \ref{subsec:nchg}), the EdgeHead, and the MSGM separately as three control-group models. As shown in Table~\ref{tab:controlgroup_corner3d}, replacing the center heatmap with the nearest corner heatmap improves the \APBEV, \APnew, and \APcs of CenterPoint by 3.5\%, 34.4\%\, and 67.2\%, with a drop of 27.7\% in \APthreeD. Such significant improvement in \APnew and \APcs highlights the contribution of the nearest corner heatmap in enhancing the detection of closer surfaces. Meanwhile, equipping CenterPoint with the EdgeHead also results in similar performance improvement, especially under \APnew and \APcs. Moreover, integrating MSGM into CenterPoint significantly improves \APBEV and \APthreeD, demonstrating its effectiveness in enhancing the model’s adaptability to cross-domain tasks. However, this adaptation also results in performance drops in \APnew and \APcs. In other words, by increasing adaptability, the model relies more on estimation during training and inference rather than learning based on available prior knowledge (\eg the point cloud data), introducing more uncertainty. Finally, by simultaneously employing the nearest corner heatmap generation module, the EdgeHead, and the MSGM, we achieve a well-balanced trade-off between the overall cross-domain detection performance of the entire 3D bounding boxes and the localization accuracy of the closer surfaces relative to the ego vehicle (\ie the LiDAR sensor), providing a more robust solution for cross-domain 3D object detection tasks.


\section{Conclusion}

In this paper, we explored more robust solutions for cross-domain 3D object detection. We first proposed two additional evaluation metrics, \ie the \APnew and \APcs, to measure 3D object detection models' ability of detecting the closer surfaces to the LiDAR sensor. These metrics can be used with the commonly used \APBEV and \APthreeD to evaluate models' cross-domain performance more comprehensively and reasonably. Furthermore, we proposed the EdgeHead, a refinement head that can be equipped with existing detectors to guide them to focus more on the closer surfaces of objects to the LiDAR sensor instead of the entire boxes. To maximize the utilization of available data relative to objects (\eg the points surrounding their surfaces), we further proposed a novel detector, CornerPoint3D, along with the nearest corner heatmap generation module and the MSGM. This anchor-free detector is designed and trained to find the nearest corners of objects and generate 3D bounding boxes based on these key points. This learning strategy, combined with the adaptability improvements brought by MSGM, enhances the model’s robustness in cross-domain tasks while maintaining the detection ability for the closer surfaces of objects. By combining the EdgeHead and CornerPoint3D, we achieve improved performance compared to the prior arts, realizing a balanced trade-off between the
detection quality of the entire bounding boxes and the locating accuracy of closer surfaces to the
LiDAR sensor. CornerPoint3D and CornerPoint3D-Edge also integrate well with ROS, the existing cross-domain data augmentation method, indicating strong compatibility with other cross-domain data augmentation methods. The success of our approach indicates the effectiveness of guiding models to learn from available data characteristics more efficiently rather than to rely on uncertain estimations or overfitting to annotations, leading to a more practical and robust cross-domain 3D object detection solution.

\backmatter

\section*{Declarations}

\begin{itemize}
\item Data availability: Datasets used in this study are publicly available. We used KITTI, Waymo, and nuScenes datasets for our study on 3D object detection task. KITTI, Waymo, and nuScenes are respectively available at:
\begin{itemize}
\item https://www.cvlibs.net/datasets/kitti/  \item https://waymo.com/open/  
\item https://www.nuscenes.org/
\end{itemize}
\end{itemize}







\begin{appendices}



\section{Box generation process in EdgeHead}\label{appendixsec:edgehead}
In the design of EdgeHead, we initially considered using ideas that generate the predictions from the box corners. However, we decided to keep the box generation process as the refinement stage of anchor-based methods due to the below two main reasons. 

Firstly, to generate boxes from the corner, we need to use the following information: (a) the coordinates of the corner; (b) the dimensions of the box, \ie the length, width, and height; and (c) the rotation angle of the box. This is similar with the required information when generating boxes based on anchors. However, the box generation from the corner requires one more input, \ie the relative position between the corner and the center of the box. As there are four corners for each box, four boxes can be generated for the same input as described above if we do not tell the relative position. This involves additional variables during the training process, making the structure of the refinement head more complex. Meanwhile, it provides more noise into the prediction of the boxes, since it is impossible that a model can 100\% correctly predict the relative position between the corner and the center. 
Secondly, the main purpose of our EdgeHead is to find an efficient and effective way to guide models to focus more on the detection quality of the closer surfaces instead of the center, we are therefore not intended to modify the structure of the existing models greatly. Using the corner-based box generation method, it is more natural to modify models from the first stage than to add an extra refinement head.

As a result, we decided to follow the anchor-based methods to generate boxes in our EdgeHead. It is also interesting to consider the corner-based box generation methods as a part of our future work.

\section{Results of EdgeHead in within-domain tasks}\label{appendixsec:withinresults}

We provide additional results of the SECOND and CenterPoint models in within-domain tasks, including KITTI $\rightarrow$ KITTI, nuScenes $\rightarrow$ nuScenes, and Waymo $\rightarrow$ Waymo, see Tables \ref{tab:within-domain-second} and \ref{tab:within-domain-centerpoint}. Note that since ROS and SN are domain adaptation methods designed for cross-domain tasks, we skip evaluating the models equipped with them in within-domain tasks. The results show that equipping with our EdgeHead can also improve the models' performance in within-domain tasks, under both the original BEV and 3D metrics and the proposed CS-ABS and CS-BEV metrics.

\begin{table}[htbp]
\caption{Performance of SECOND under four metrics in within-domain tasks.}
\label{tab:within-domain-second}
\footnotesize
\setlength\tabcolsep{2pt}
\centering
\begin{tabular}{c c c c c c} 
\toprule
Task & Method & \APBEV / \APthreeD & \APnew / \APcs \\
\midrule
\multirow{2}{*}{K $\rightarrow$ K}  
& Original & 84.3 / 72.1 & 71.4 / 53.3 \\
& + EdgeHead & 88.2 / 77.5 & 79.6 / 66.7 \\
\midrule
\multirow{2}{*}{N $\rightarrow$ N}    
& Original & 43.1 / 22.6 & 25.9 / 14.2\\
& + EdgeHead & 46.7 / 29.7 & 34.6 / 25.2 \\
\midrule
\multirow{2}{*}{W $\rightarrow$ W} 
& Original & 63.1 / 47.5 & 47.3 / 34.2 \\
& + EdgeHead & 65.1 / 51.4 & 53.8 / 44.1 \\
\bottomrule
\end{tabular}
\end{table}

\begin{table}[htbp]
\caption{Performance of CenterPoint under four metrics in within-domain tasks.}
\label{tab:within-domain-centerpoint}
\footnotesize
\setlength\tabcolsep{2pt}
\centering
\begin{tabular}{c c c c c c} 
\toprule
Task & Method & \APBEV / \APthreeD & \APnew / \APcs \\
\midrule
\multirow{2}{*}{K $\rightarrow$ K}  
& Original & 84.4 / 73.6 & 72.8 / 56.8 \\
& + EdgeHead & 84.4 / 75.3 & 74.7 / 61.8 \\
\midrule
\multirow{2}{*}{N $\rightarrow$ N}    
& Original & 49.7 / 30.7 & 34.1 / 23.2 \\
& + EdgeHead & 48.1 / 32.5 & 36.9 / 26.9 \\
\midrule
\multirow{2}{*}{W $\rightarrow$ W} 
& Original & 67.3 / 52.5 & 52.6 / 41.4 \\
& + EdgeHead & 65.4 / 53.8 & 56.0 / 46.7 \\
\bottomrule
\end{tabular}
\end{table}

\section{Preliminaries for CornerPoint3D}\label{sec:preliminaties_cornerpoint3d}

\noindent\textbf{CenterPoint~\cite{Yin_2021_CVPR}} predicts the centers of objects in heatmap representations and generates entire 3D bounding boxes based on the top-ranking centers. Following the idea in CenterNet~\cite{zhou2019objects}, multiple heatmaps $\hat{Y}$ are generated to independently represent the potential center locations for each object class. A focal loss~\cite{lin2017focal} is applied to supervise the quality of heatmaps. All the potential center locations $c = (c_x, c_y) \in \mathbb{R}^{2}$ are splatted onto the heatmap of the relevant class by using a modified Gaussian kernel function, \ie 
\begin{equation}
    Y_{xy} = \exp\left(-\frac{(x - c_x)^2 + (y - c_y)^2}{2\sigma_c^2}\right),
\end{equation}
where $\sigma_c = \max(f(wl), \tau)$. The function $f$ is an object size-adaptive function to calculate the radius defined in CornerNet~\cite{Law_2018_ECCV}, and $\tau = 2$ is the smallest allowable Gaussian radius defined by CenterPoint for denser positive supervisory signals~\cite{Yin_2021_CVPR}. A maximum value will be taken if two or more Gaussians overlap at some positions. 

\begin{figure}[htbp]
\begin{center}
   \includegraphics[width=\linewidth]{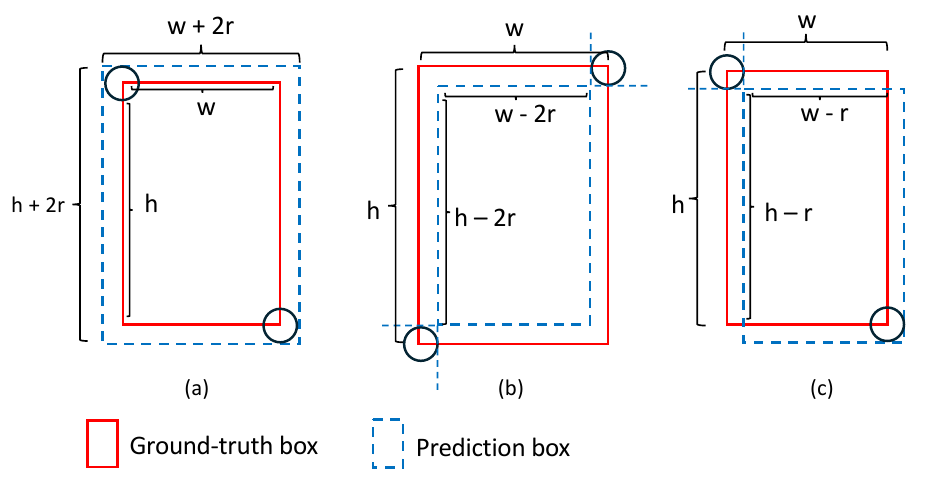}
\end{center}
   \caption[Three conditions of radius calculation in CornerNet]{Three conditions of radius calculation in CornerNet~\cite{Law_2018_ECCV}. (a) The top-left and bottom-right predicted corners are both outside of the ground-truth box. (b) The top-left and bottom-right corners are both inside of the ground-truth box. (c) One of the corners is outside, and the other is inside the ground-truth box.}
\label{fig:cornernet_radius}
\end{figure}

\noindent\textbf{Radius calculation in CornerNet.} In detail, the radius $r$ (with abuse of notation) of the Gaussian kernel function for the corners in CornerNet~\cite{Law_2018_ECCV} is decided based on the size of the related object by ensuring that a pair of points within the radius would generate a bounding box whose IoU with the ground-truth box is larger than a predetermined threshold. As shown in Figure~\ref{fig:cornernet_radius}, the relationship between the predicted corner and the ground-truth corner can be divided into three cases, \ie (a) the top-left and bottom-right predicted corners are both outside of the ground-truth box; (b) the top-left and bottom-right corners are both inside of the ground-truth box; and (c) one of the corners is outside, and the other one is inside the ground-truth box. Given the height $h$ and width $w$ of the (2D) bounding box and the radius $r$, we can calculate the relevant overlap $o$ of the predicted box and the ground-truth box for the three cases as below:
\begin{align}
    &o = \frac{h  w}{(h+2r) (w+2r)}, \\
    &o = \frac{(h - 2r) (w - 2r)}{hw }, \\
    &o = \frac{(h - r) (w - r)}{2 h w - (h - r) (w - r)}.
\end{align}
Afterwards, by transforming the equations into the standard format of quadratic equations in terms of $r$, we can solve the above three equations and select the minimum solution as the final radius of the Gaussian radius. 

\section{Comparison between CornerPoint3D and SECOND}\label{sec:additonal_results_second}

We provide additional results comparison between our method CornerPoint3D and the SECOND~\cite{s18103337} in Table \ref{tab:cornerpoint3d_with_edgehead_second}. Analogous to the format in Table~\ref{tab:cornerpoint3d_with_edgehead} in the main paper, in Table \ref{tab:cornerpoint3d_with_edgehead_second}, we first compare the results of SECOND with CornerPoint3D, and then apply the EdgeHead~\cite{zhang2024detect} to SECOND to compare the performance with CornerPoint3D-Edge. Afterwards, we apply ROS to SECOND and CornerPoint3D. Last but not least, we apply ROS and EdgeHead to SECOND at the same time, and compare with CornerPoint3D-Edge equipped with ROS.

\begin{table*}[htbp]
    \caption[Main results of CornerPoint3D and CornerPoint3D-Edge, and comparisons with SECOND under different tasks]{Main results of CornerPoint3D and CornerPoint3D-Edge, and comparisons with SECOND~\cite{s18103337} under different tasks. We report \APBEV,  \APthreeD and \APcs of the car category at IoU = 0.7 and \APnew at IoU = 0.5. The reported performance is the moderate case when KITTI is the target domain, and is the overall result for other cross-domain tasks. Improvements are calculated by the relative difference between each SECOND and CornerPoint3D equipped with the same methods (\eg native, the EdgeHead (Edge)~\cite{zhang2024detect}, and the ROS~\cite{wang2020train}).  W, K, and N represent the Waymo, KITTI, and nuScenes datasets, respectively. }
    \label{tab:cornerpoint3d_with_edgehead_second}
    \small
    \setlength\tabcolsep{4pt}
    \centering
    \resizebox{0.96\textwidth}{!}{
    \begin{tabular}{l l l l l l}
    \toprule[1pt]
    Task & Method & \APBEV & \APthreeD & \APnew & \APcs \\
    \midrule
    \multirow{8}{*}{W $\rightarrow$ K} 
    & SECOND & 49.2 & 9.3 & 19.0 & 10.9  \\
    & \textbf{CornerPoint3D} & \textbf{47.5} (-3.5\%) & \textbf{8.4} (-9.7\%) & \textbf{20.0} (+5.3\%) & \textbf{11.6} (+6.4\%) \\
    & SECOND w/ Edge & 52.3 & 10.7 & 23.7 & 14.7 \\
    & \textbf{CornerPoint3D-Edge} & \textbf{58.9} (+12.6\%) & \textbf{12.4} (+15.9\%) & \textbf{28.3} (+19.4\%) & \textbf{18.6} (+26.5\%) \\
    & SECOND w/ ROS & 73.0 & 38.3 & 33.7 & 12.6 \\
    & \textbf{CornerPoint3D w/ ROS} & \textbf{70.0} (-4.1\%) & \textbf{29.9} (-21.9\%) & \textbf{39.2} (+16.3\%) & \textbf{19.3} (+53.2\%) \\
    & SECOND w/ Edge \& ROS & 76.4 & 41.5 & 42.9 & 20.4 \\
    & \textbf{CornerPoint3D-Edge w/ ROS} & \textbf{80.7} (+5.6\%) & \textbf{48.7} (+17.3\%) & \textbf{50.9} (+18.7\%) & \textbf{24.8} (+21.6\%) \\
    \midrule
    \multirow{8}{*}{W $\rightarrow$ N} 
    & SECOND & 27.8 & 16.1 & 15.6 & 6.7  \\
    & \textbf{CornerPoint3D} & \textbf{25.8} (-7.2\%) & \textbf{11.6} (-28.0\%) & \textbf{15.9} (+1.9\%) & \textbf{9.1} (+35.8\%) \\
    & SECOND w/ Edge & 29.9 & 18.0 & 20.9 & 13.0 \\
    & \textbf{CornerPoint3D-Edge} & \textbf{29.8} (-0.3\%) & \textbf{15.0} (-16.7\%) & \textbf{21.2} (+1.4\%) & \textbf{13.0} (0.0\%) \\
    & SECOND w/ ROS & 26.7 & 15.4 & 15.8 & 6.5 \\
    & \textbf{CornerPoint3D w/ ROS} & \textbf{24.7} (-7.5\%) & \textbf{10.2} (-33.8\%) & \textbf{14.5} (-8.2\%) & \textbf{7.7} (+18.5\%) \\
    & SECOND w/ Edge \& ROS & 28.3 & 17.1 & 19.9 & 11.9 \\
    & \textbf{CornerPoint3D-Edge w/ ROS} & \textbf{29.7} (+4.9\%) & \textbf{15.3} (-10.5\%) & \textbf{21.2} (+6.5\%) & \textbf{13.3} (+11.8\%) \\
    \midrule
    \multirow{8}{*}{N $\rightarrow$ K} 
    & SECOND & 35.7 & 11.8 & 16.4 & 9.8  \\
    & \textbf{CornerPoint3D} & \textbf{43.0} (+20.4\%) & \textbf{12.5} (+5.9\%) & \textbf{24.1} (+47.0\%) & \textbf{12.9} (+31.6\%) \\
    & SECOND w/ Edge & 53.6 & 15.9 & 33.3 & 19.6 \\
    & \textbf{CornerPoint3D-Edge} & \textbf{49.7} (-7.3\%) & \textbf{18.5} (+16.4\%) & \textbf{32.5} (-2.4\%) & \textbf{21.9} (+11.7\%) \\
    & SECOND w/ ROS & 43.4 & 20.0 &  20.2 & 8.1 \\
    & \textbf{CornerPoint3D w/ ROS} & \textbf{42.6} (-1.8\%) & \textbf{20.1} (+0.5\%) & \textbf{30.9} (+52.9\%) & \textbf{19.2} (+137.0\%) \\
    & SECOND w/ Edge \& ROS & 52.7 & 33.1 & 39.9 & 24.6 \\
    & \textbf{CornerPoint3D-Edge w/ ROS} & \textbf{65.3} (+23.8\%) & \textbf{41.9} (+26.6\%) & \textbf{53.2} (+33.3\%) & \textbf{35.4} (+43.9\%) \\
    \bottomrule[0.8pt]
\end{tabular}
        }
\end{table*}

\section{More heatmap visualizations}\label{sec:additonal_heatmaps}

Following our visualizations in Figure~\ref{fig:heatmaps_w2k} in the main paper, we provide more visualizations of the nearest corner heatmaps from the CornerPoint3D, and compare them with the center heatmaps from the CenterPoint~\cite{Yin_2021_CVPR}. We use the models trained on Waymo and nuScenes respectively, and test them on the KITTI dataset.

\begin{figure*}[htbp]
\begin{center}
   \includegraphics[width=0.9\linewidth]{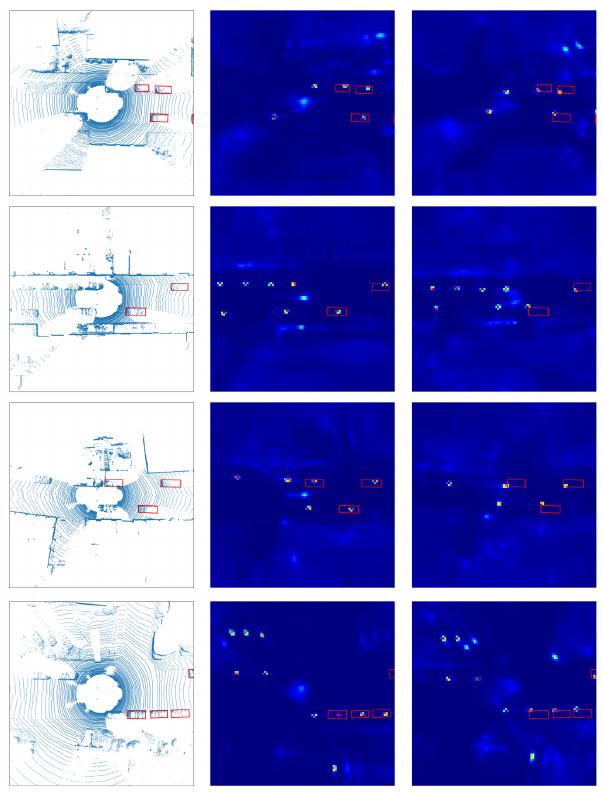}
   \put(-384,545){{\small Original Point Cloud}}
   \put(-230,545){{\small CenterPoint}}
   \put(-118,545){{\small \textbf{CornerPoint3D (Ours)}}}
   \put(-425,445){\rotatebox{90}{\small Sample ID - 000031}}
   \put(-425,305){\rotatebox{90}{\small Sample ID - 000039}}
   \put(-425,170){\rotatebox{90}{\small Sample ID - 000050}}
   \put(-425,35){\rotatebox{90}{\small Sample ID - 000093}}
\end{center}
   \caption[Nearest corner heatmaps and center heatmaps from CornerPoint3D and CenterPoint respectively]{Comparison between the nearest corner heatmaps (right column) from CornerPoint3D (ours) and the center heatmaps (middle column) from CenterPoint~\cite{Yin_2021_CVPR}. Visualizations of the original point cloud data are also provided (left column), magnified by 400\%. Both models are for the Waymo $\rightarrow$ KITTI task (\ie trained on Waymo and tested on KITTI).}
\label{additional_fig:heatmaps_w2k_1}
\end{figure*}

\begin{figure*}[htbp]
\begin{center}
   \includegraphics[width=0.9\linewidth]{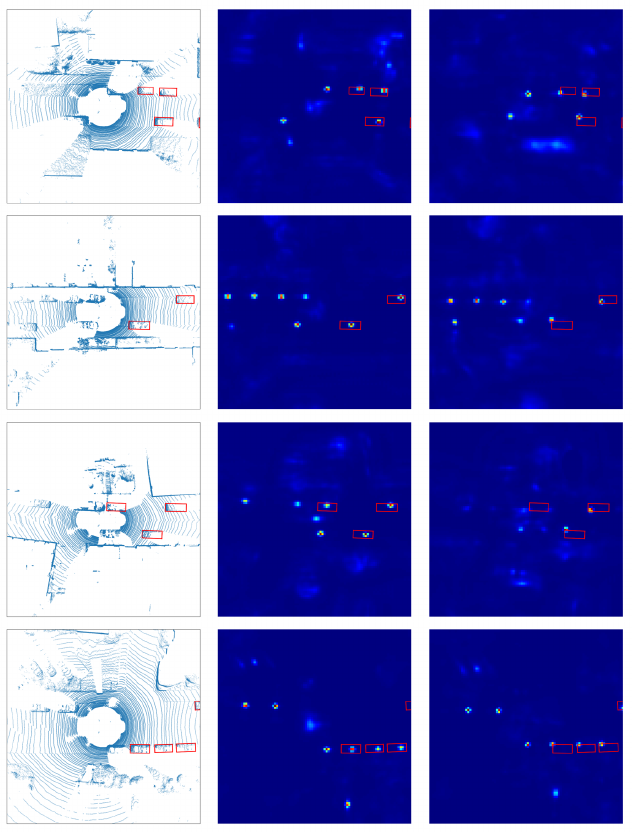}
   \put(-384,545){{\small Original Point Cloud}}
   \put(-230,545){{\small CenterPoint}}
   \put(-118,545){{\small \textbf{CornerPoint3D (Ours)}}}
   \put(-425,445){\rotatebox{90}{\small Sample ID - 000031}}
   \put(-425,305){\rotatebox{90}{\small Sample ID - 000039}}
   \put(-425,170){\rotatebox{90}{\small Sample ID - 000050}}
   \put(-425,35){\rotatebox{90}{\small Sample ID - 000093}}
\end{center}
   \caption[Nearest corner heatmaps and center heatmaps from CornerPoint3D and CenterPoint respectively]{Comparison between the nearest corner heatmaps (right column) from CornerPoint3D (ours) and the center heatmaps (middle column) from CenterPoint~\cite{Yin_2021_CVPR}. Visualizations of the original point cloud data are also provided (left column), magnified by 400\%. Both models are for the nuScenes $\rightarrow$ KITTI task (\ie trained on nuScenes and tested on KITTI).}
\label{additional_fig:heatmaps_n2k_2}
\end{figure*}




\end{appendices}



\bibliography{sn-bibliography}

\end{document}